\documentclass[letterpaper]{article} 
\usepackage[preprint]{aaai2027}  
\usepackage[hyphens]{url}  
\usepackage{graphicx} 
\usepackage{natbib}  
\usepackage{caption} 
\usepackage{algorithm}
\usepackage{algorithmic}

\usepackage{newfloat}
\usepackage{listings}
\DeclareCaptionStyle{ruled}{labelfont=normalfont,labelsep=colon,strut=off} 
\floatstyle{ruled}
\newfloat{listing}{tb}{lst}{}
\floatname{listing}{Listing}

\usepackage{booktabs}

\usepackage{amssymb}
\usepackage{amsmath}
\usepackage{multirow}

\title{The K-Space Signature: Frequency-Domain Representation Learning for Medical Deepfake Detection}
\author {
    Riccardo Raciti\textsuperscript{\rm 1},
    Francesco Guarnera\textsuperscript{\rm 1},
    Francesco Rundo\textsuperscript{\rm 1},
    Luca Guarnera\textsuperscript{\rm 1}\thanks{These authors contributed equally as joint senior authors.},
    Sebastiano Battiato\textsuperscript{\rm 1}\footnotemark[1]
}
\affiliations {
    \textsuperscript{\rm 1}University of Catania\\
    riccardo.raciti@phd.unict.it, \{francesco.guarnera, francesco.rundo, luca.guarnera, sebastiano.battiato\}@unict.it
}

\begin{document}

\maketitle

\begin{abstract}
In medical imaging, generative models are increasingly deployed to synthesize realistic data and augment limited datasets. Unfortunately, while beneficial for privacy-preserving data sharing, these synthesized images can be repurposed for malicious intents, threatening public health through the creation of Medical Deepfakes. To address this threat, we introduce the K-Space Signature (KSS), a novel forensic framework that isolates hardware and generative traces within the spectral domain. By shifting analysis to the frequency domain, the KSS suppresses macroscopic anatomical variance by subtracting an empirical global anatomical prior computed in the Logarithmic Power Spectral Density (Log-PSD) space. To effectively process these globally distributed spectral artifacts without the local spatial bias inherent to Convolutional Neural Networks, we pair the KSS representation with a novel 3D MLP-Mixer architecture equipped with an ArcFace metric-learning head. Extensive experiments on multi-center 3D MRI datasets demonstrate that this combined approach achieves exceptional detection performance, exceeding 0.99 Accuracy and ROC-AUC on multi-generator synthetic datasets. Furthermore, the framework exhibits robust zero-shot generalization, maintaining strong discriminative power (up to 0.93 Accuracy) on independent datasets acquired from entirely unseen scanners. To ensure full reproducibility, the complete source code and pre-trained models will be made publicly available upon acceptance.
\end{abstract}

\begin{links}
    \link{Code}{https://github.com/Raciti/KSS.git}
\end{links}

\section{Introduction}
The rapid evolution of generative models, such as MAISI-v2~\cite{maisi} and Med-DDPM~\cite{med-ddpm}, has enabled the synthesis of highly realistic medical data, including 3D Magnetic Resonance Imaging (MRI) volumes~\cite{maisi,med-ddpm}. While beneficial for data augmentation and privacy-preserving data sharing, these advancements introduce severe vulnerabilities regarding data provenance, security, and clinical integrity~\cite{char2018implementing}. Although the susceptibility of general computer vision models to malicious manipulation is widely recognized, medical image analysis systems exhibit unique vulnerabilities due to the specialized nature of clinical diagnostic workflows~\cite{bortsova2021adversarial,finlayson2019adversarial,tsai2023adversarial}. Consequently, the introduction of fully synthetic scans, \textit{Medical Deepfakes}, into healthcare systems poses an unprecedented threat, making automatic authenticity verification essential. Medical Deepfakes threaten diagnostic integrity by seamlessly deceiving standard detection systems~\cite{li2025toward}.
While various forensic methods have been extensively developed to identify deepfakes~\cite{yermakov2026deepfake,guarnera2024mastering,siddiqui2025enhanced,yang2025d,guarnera2020deepfake}, these techniques are not directly applicable to the medical domain. This limitation arises because the structural formats of MRI volumes differ fundamentally from those of standard 2D images, videos, and audio signals. Medical MRI scanners inherently embed unique, acquisition-specific background noise and hardware traces~\cite{kushol2023effects}. However, general forensic methods struggle to disentangle these physical hardware traces from learned algorithmic artifacts, particularly failing to isolate globally distributed anomalies in the frequency domain. Although these intrinsic hardware traces are implicitly exploited for scanner model and manufacturer identification~\cite{fang2020deep, mohammadi2022deep}, repurposing them for deepfake detection remains highly challenging. Recent advancements have attempted to bridge this gap by relying on cross-domain feature extraction~\cite{li2025toward} or utilizing unsupervised backward diffusion to detect subtle disruptions in natural background noise~\cite{grabovski2025back}. Despite these efforts, existing frameworks still fail to explicitly isolate these hardware-specific signatures within the frequency domain for robust Medical Deepfake detection.
\begin{figure*}[!t]
    \centering
    \includegraphics[width=\textwidth]{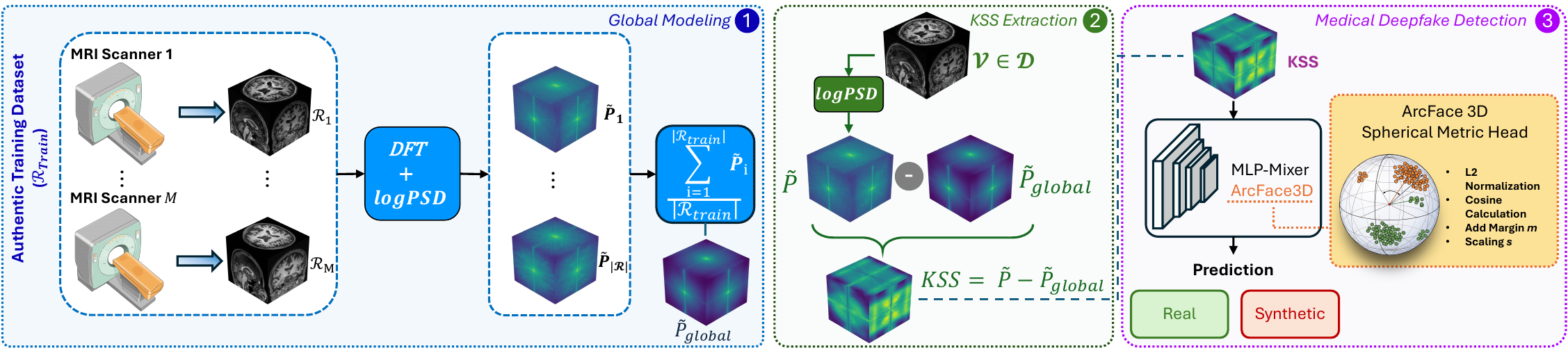} 
    \caption{Overview of the proposed framework. (1) Global Modeling: A universal anatomical prior, $\tilde{P}_{global}$, is computed as the empirical mean of authentic LogPSD MRI volumes $(\mathcal{R}_{train})$. (2) KSS Extraction: For an input volume with LogPSD $\tilde{P}$, subtracting $\tilde{P}_{global}$ isolates non-anatomical hardware and generative artifacts into the KSS residual. (3) Medical Deepfake: An MLP-Mixer and ArcFace 3D Metric Head apply $L_{2}$ normalization, an angular margin $(m)$, and scaling $(s)$ to the KSS tensor, explicitly separating physical signatures from synthetic algorithmic traces for binary classification.}
    \label{fig:pipeline}
\end{figure*}
Inspired by photo-response non-uniformity (PRNU) analysis for camera source attribution~\cite{lukas2006digital}, we hypothesize that authentic MRI volumes contain intrinsic scanner-specific acquisition traces, whereas synthetic volumes are characterized by generative signatures~\cite{marra2019gans}. Generative architectures inherently struggle to faithfully replicate the complex physical noise of clinical MR scanners; consequently, analyzing this spectral domain naturally exposes the fundamental discrepancies between authentic and synthetic volumes. To explicitly capture these discrepancies, we introduce the KSS, a novel representation obtained by mapping 3D MRI volumes to the frequency domain and projecting them into a LogPSD space. By subtracting a dynamically computed global anatomical baseline, we filter out dominant low-frequency macroscopic anatomy. The resulting KSS tensor acts as a representation of non-anatomical traces that serves a dual purpose: enabling fine-grained scanner model classification and providing a hardware-invariant feature space for robust real-vs-fake detection. To effectively process these globally distributed frequency-domain signatures without spatial bias, we pair this representation with a novel 3D MLP-Mixer architecture equipped with an ArcFace metric-learning head. Fig.~\ref{fig:pipeline} illustrates the proposed framework, from KSS extraction to classification.
\\
\textbf{Contributions} To the best of our knowledge, this is the first work that isolates and analyzes the scanner and generative traces in the frequency domain for Medical Deepfake Detection. Specifically, the main contributions are fivefold: \textbf{(1)} we formalize a method to isolate and extract hardware-specific traces from 3D MRI volumes, moving beyond black-box empirical observations; \textbf{(2)} we validate the KSS domain as a discriminative representation space for medical image forensics, demonstrating that spectral features restrict models from relying on spatial anatomical shortcuts; \textbf{(3)} we demonstrate that these isolated signatures capture distinct, manufacturer-specific hardware topologies, revealing strong intra-vendor correlation and mathematically orthogonal inter-vendor domains; \textbf{(4)} we introduce the MLP-Mixer3D ArcFace architecture, a novel deep learning framework specifically designed for the spectral domain, combining spatial-token mixing with additive angular margin constraints; and \textbf{(5)} we combine the KSS and MLP-Mixer3D ArcFace to effectively differentiate authentic clinical volumes from state-of-the-art synthetic generations.

\begin{figure*}[!ht] 
    \centering
    \includegraphics[width=\linewidth]{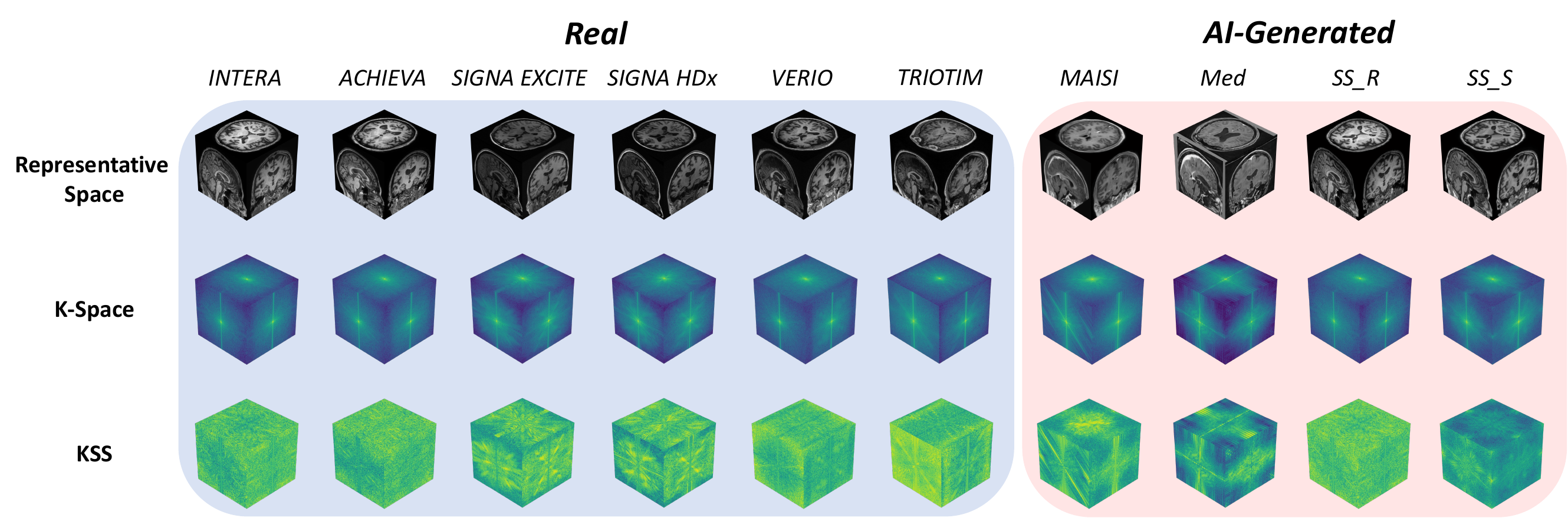} 
    \caption{Real and synthetic MRI across domains. Rows: spatial anatomy (top), K-Space (middle), and KSS traces (bottom). Left: real scans from six scanners (Intera, Achieva, SIGNA EXCITE, SIGNA HDx, Verio, TrioTim). Right: generative signatures from MAISI, Med, and SuperSynth (SS\_R, resampled; SS\_S, synthetic contrast).}
    \label{fig:dataset}
\end{figure*}

\section{Methodology}
In this section, we detail the mathematical foundations of our framework, formalize the proposed KSS isolation, and present the theoretical design of the architecture.

\subsection{Problem Formulation}
Let $\mathcal{V} \subset \mathbb{R}^{D \times H \times W}$ denote the space of 3D MRI volumes. We formalize our evaluation dataset as $\mathcal{D} = \{(V_i, y_i) \;|\; V_i \in \mathcal{V}\}_{i=1}^{|\mathcal{D}|}$, where $V_i$ is an individual volume and $y_i$ is its corresponding class label. The dataset is partitioned into $\mathcal{D} = \mathcal{R} \cup \mathcal{S}$ where $\mathcal{R} , \mathcal{S} \subset \mathcal{V}$ represent the sets of authentic and synthetic MRI, respectively, with $\mathcal{R} \cap \mathcal{S} = \emptyset$. To model multi-center hardware heterogeneity, the authentic set is aggregated from $M$ physical MRI scanner models $\mathcal{M} = \{\mathcal{M}_i\}_{i=1}^M$, such that $\mathcal{R} = \bigcup_{i=1}^M \mathcal{R}_i$. Conversely, the synthetic MRI are produced by a set of $Q$ distinct generative pipelines $\mathcal{G} = \{\mathcal{G}_k\}_{k=1}^Q$, yielding a total synthetic set $\mathcal{S} = \bigcup_{k=1}^Q \mathcal{S}_k$, where $\mathcal{S}_k$ denotes the subset of MRI generated by pipeline $\mathcal{G}_k$.

To rigorously prevent anatomical data leakage during evaluation, we partition $\mathcal{D}$ into training, validation, and test subsets (80:10:10 distribution) strictly at the subject level. Let $\mathcal{U}$ denote the set of all unique patient identities, and $u: \mathcal{V} \rightarrow \mathcal{U}$ define the mapping from any given volume to its corresponding patient. We first partition the patient pool into mutually exclusive and exhaustive subsets $\mathcal{U}_{\tau}$ for $\tau \in \{train, val, test\}$. The corresponding dataset subsets are then formally defined as:

\begin{equation}
    \begin{split}
        \mathcal{X}_{\tau} &= \{v \in \mathcal{X} \mid u(v) \in \mathcal{U}_{\tau}\}, \\
        &\quad \text{for } \mathcal{X} \in \{\mathcal{R}, \mathcal{S}\} \\
        &\quad \text{and } \tau \in \{\text{train, val, test}\}
    \end{split}
\end{equation}

This formulation guarantees that all volumes belonging to a specific patient, regardless of the scanner model or generative pipeline, are strictly isolated to a single subset. This intrinsically satisfies the exhaustiveness and disjointness constraints ($\mathcal{X}_a \cap \mathcal{X}_b = \emptyset$ for $a \neq b$). Finally, patient assignments to $\mathcal{U}_{\tau}$ are stratified to ensure each authentic split $\mathcal{R}_{\tau}$ and synthetic split $\mathcal{S}_{\tau}$ maintains a balanced representation of the $M$ hardware domains and $Q$ generative pipelines, respectively.
While spatial volumes are essential for clinical diagnostics, learning directly within $\mathcal{V}$ makes models highly prone to overfitting patient-specific anatomical semantics rather than fundamental acquisition artifacts. 
\paragraph{Frequency Domain Mapping.} While raw $k$-space represents the initial complex data acquired by the MRI hardware, applying a 3D Discrete Fourier Transform (DFT) to a fully reconstructed spatial volume provides a mathematically equivalent frequency-domain representation~\cite{gallagher2008introduction,kopanoglu2015radiofrequency}, for the remainder of this work, we will refer to this DFT-derived frequency-domain representation simply as K-space. This transition to the frequency domain is motivated by recent findings in medical image processing and synthetic forensics, which demonstrate that spectral representations effectively mitigate spatial overfitting~\cite{tan2024frequency, li2023zero} while exposing subtle, high-frequency hardware and generative artifacts~\cite{gao2024deepfake, mejri2021leveraging, giudice2021fighting}. For any individual volume $V \in \mathcal{V}$, its complex-valued frequency representation $\hat{V} \in \mathbb{C}^{D \times H \times W}$ is explicitly computed via the DFT, denoted as $\mathcal{F}$, such that $\hat{V} = \mathcal{F}(V)$. Specifically, for any spatial coordinate $(x, y, z)$ and corresponding frequency coordinate $(k_x, k_y, k_z)$, the complex-valued frequency representation $\hat{V}$ is explicitly computed as:
\begin{equation}
\begin{split}
\hat{V}(k_x, k_y, k_z) &= \sum_{x=0}^{D-1} \sum_{y=0}^{H-1} \sum_{z=0}^{W-1} V(x, y, z) \\
&\quad \exp\left( -j 2\pi \left( \frac{k_x x}{D} + \frac{k_y y}{H} + \frac{k_z z}{W} \right) \right)
\end{split}
\end{equation}

where $j$ represents the imaginary unit. To analyze the energy distribution of the signal, we compute the LogPSD, referred to as $\tilde{P}$:
\begin{equation}
    \tilde{P} = \ln(1 + |\hat{V}|^{2})
    \label{eq:LogPSD}
\end{equation}
where $|\cdot|^{2}$ represents the element-wise squared absolute value.
The resulting $\tilde{P}$ compresses the dynamic range, ensuring stable numerical behavior and appropriately weighting both macro-structures (low frequencies) and subtle hardware or generative artifacts, which become more prominent in the high-frequency spectrum. However, $\tilde{P}$ remains largely dominated by patient-specific anatomical content, motivating the need for a more discriminative forensic representation.

\paragraph{K-Space Signature (KSS).} To isolate the non-anatomical hardware and algorithmic traces, we define a transition from the standard K-space to the isolated 
\textit{KSS} domain. 

We hypothesize that the aggregate spatial frequency distribution over a sufficiently large and diverse dataset converges to a global anatomical prior, which characterizes the intrinsic spatial frequency priors of the human brain. We define this baseline, $\tilde{P}_{\text{global}}$, as the empirical mean of $\tilde{P}$ computed exclusively over the training partition of the authentic dataset $\mathcal{R}_{\text{train}}$:
\begin{equation}
\tilde{P}_{global} = \frac{1}{|\mathcal{R}_{train}|} \sum_{V_i \in \mathcal{R}_{train}}\ln(1 + |\hat{V}|^{2})
\label{global_formula}
\end{equation}
Computing $\tilde{P}_{global}$ strictly on authentic training volumes ensures two critical properties: first, it prevents downstream data leakage from the validation and testing sets; second, it guarantees that the baseline models a isolated anatomical frequency distribution entirely uncontaminated by the synthetic artifacts or algorithmic signatures present in $\hat{V}$. The $\tilde{P}_{global}$ averages out idiosyncratic patient differences and scanner-specific biases; it isolates the intrinsic anatomical frequency decay shared across human neuroanatomy. We validate the robustness of this global anatomical prior assumption across varying distributions in the Supplementary Material.

To disentangle the underlying anatomical structure from the hardware trace (e.g., characteristic noise, coil sensitivities, and filtering) \cite{zuo2021unsupervised, chartsias2019disentangled}, we subtract the global anatomical baseline, $\tilde{P}_{global}$, ffrom the $k$-space representation of the $i$-th volume. We formalize this residual as the isolated K-Space Signature $KSS_i$ for that specific volume:
\begin{equation}
    KSS_i = \tilde{P}_i - \tilde{P}_{global}
    \label{eq:kss}
\end{equation} 

The resulting tensor, $KSS_{i}$, is a scanner-associated representation of the artifacts and hardware prints unique to that exact scan, independent of the underlying patient anatomy. Fig.~\ref{fig:dataset} shows visual examples of these isolated signatures across both authentic scanner models and synthetic generators.

In the Supplementary material, we demonstrate the robustness and consistency of $\tilde{P}_{global}$ by evaluating the subtraction across varying training subsets and validating the extracted traces through scanner classification.

\subsection{MLP-Mixer3D ArcFace}
To process the high-dimensional frequency representations, we depart from traditional Convolutional Neural Networks (CNNs), which possess an inherent inductive bias toward local spatial textures. Instead, we propose the MLP-Mixer3D ArcFace architecture. While the original MLP-Mixer~\cite{tolstikhin2021mlp} was exclusively designed to process 2D planar images, we generalize its token-mixing paradigm to accommodate volumetric 3D MRI data, equipping it with a dynamic spatial padding strategy and a robust metric-learning head to capture global spectral anomalies.

\paragraph{3D Dynamic Patch Extraction.} 
Let the input KSS tensor be defined as $X \in \mathbb{R}^{D \times H \times W \times C}$, where $C=1$ represents the single-channel Log-PSD domain. Clinical 3D MRI volumes are rarely cleanly divisible by standard patch sizes, which historically leads to dimensional mismatch or restrictive cropping. To resolve this, our 3D adaptation implements a dynamic asymmetric padding strategy, appending zero-value elements to pad the spatial boundaries to the nearest optimal multiples of the patch size $P$. The padded volume, denoted as $X_{pad} \in \mathbb{R}^{D' \times H' \times W' \times C}$, allows for strict, non-overlapping extraction.

The volume is subsequently processed by a 3D convolution operator utilizing a kernel size and stride equal to $P=16$. This extracts a sequence of $P \times P \times P$ volumetric patches and linearly projects them into a hidden embedding dimension $C_e=128$. The tensor is then flattened to form the initial input matrix $Z_0 \in \mathbb{R}^{L \times C_e}$, where the total sequence length of the tokens is defined as $L = \frac{D' \cdot H' \cdot W'}{P^3}$.

\paragraph{3D Token and Channel Mixing.} The core of the network consists of $d=6$ consecutive Mixer Blocks. To successfully capture the diffuse, globally distributed spectral artifacts inherent to the frequency domain, each $l$-th block (where $l \in [1, d]$) alternates between token-mixing and channel-mixing. The operations for a single block are mathematically formalized as:

\begin{equation}
    U_l = Z_{l-1} + \text{MLP}_{token}(\text{LN}(Z_{l-1})^T)^T
\end{equation}

\begin{equation}
    Z_l = U_l + \text{MLP}_{channel}(\text{LN}(U_l))
\end{equation}

where $U_l$ denotes the intermediate matrix produced after the token-mixing stage, prior to the channel-mixing stage and $\text{LN}(\cdot)$ denotes Layer Normalization. The transpose operation $(\cdot)^T$ applied during the token-mixing stage is a critical mechanism; it permutes the axes so that $\text{MLP}_{token}$ (configured with a hidden dimension of 256) acts across the entire spatial sequence length $L$, mapping global structural correlations. Conversely, $\text{MLP}_{channel}$ (configured with a hidden dimension of 512) operates on the embedding dimension $C_e$, mapping localized feature dependencies. Both Multi-Layer Perceptrons utilize GELU non-linearities and a dropout rate of 0.4 to optimize gradient flow and prevent overfitting. After the final block, the sequence is normalized and compressed via Global Average Pooling to yield a 1D feature representation $z \in \mathbb{R}^{C_e}$.

\paragraph{ArcFace 3D Spherical Metric Head.} To effectively isolate the subtle discrepancies between multi-center scanner hardware and state-of-the-art generative pipelines, we replace the traditional Softmax classifier with an ArcFace Metric Head~\cite{deng2019arcface}. The 1D feature representation $z$ and the fully connected class weight vectors $W \in \mathbb{R}^{K \times C_e}$ (where $K$ represents the total number of evaluation classes) are strictly $L_2$-normalized to project the embeddings onto a hypersphere. 

By imposing an additive angular margin penalty $m=0.5$, we enforce rigorous inter-class separation, scaling the resulting similarity logits by a hypersphere radius scalar $s=30$. This objective formulation forces the model to map genuine physical signatures and synthetic algorithmic traces into distinct, non-overlapping angular clusters:

\begin{equation}
    \mathcal{L}_{ArcFace} = -\frac{1}{B} \sum_{i=1}^{B} \log \frac{e^{s(\cos(\theta_{y_i} + m))}}{e^{s(\cos(\theta_{y_i} + m))} + \sum_{j=1, j \neq y_i}^{K} e^{s \cos(\theta_j)}}
\end{equation}

where $B$ is the batch size, $\theta_{y_i}$ is the angle between the normalized embedding and its ground-truth class centroid, and $\theta_j$ is the angle relative to the $j$-th class. During optimization, this geometric margin is applied exclusively to the true class utilizing one-hot encoded scattering, significantly enhancing the network's discriminative power and zero-shot generalization capabilities across unseen scanners.

\section{Experiments}
\subsection{Experimental Setup} \label{exp}
\paragraph{Datasets.} We evaluate our framework across three data cohorts. \textbf{Real Dataset:} 1,200 T1-weighted scans from ADNI and PPMI, balanced across three manufacturers (GE, Philips, Siemens) and six scanner models (GE SIGNA EXCITE/HDx, Philips Achieva/Intera, Siemens Verio/TrioTim). \textbf{Synthetic Dataset:} Volumes generated by MAISI-v2 (MAISI), Med-DDPM (Med), and SuperSynth (providing resampled geometry, SS\_R, and synthetic contrast, SS\_S)~\cite{billot2023synthseg}. \textbf{Open-Set Dataset:} 120 authentic scans from unseen models (GENESIS\_SIGNA, MR 7700, Symphony). To rigorously assess out-of-distribution generalization, we evaluate against only one synthetic dataset at a time during testing.

All volumes underwent uniform spatial standardization to the MNI152 template. Comprehensive details regarding patient demographics, specific scanner hardware, synthetic generation pipelines and preprocessing interpolation are provided in the Supplementary Materials.

\paragraph{Evaluation Protocol.}
We formulate detection as a binary classification task, authentic clinical scans versus synthetic generations, evaluated via Accuracy, PR-AUC, and ROC-AUC. To assess zero-shot domain adaptation and robustness against hardware-induced shifts, we employ a Leave-One-Domain-Out validation strategy across scanner manufacturers. Combined with strict patient-level isolation, this rigorous setup forces the framework to learn universal, scanner-invariant generative artifacts rather than overfitting to specific hardware noise profiles or anatomical shortcuts.

\subsection{Experiments and Results}
We evaluate our approach across four primary dimensions, ordered as follows: (1) the structural isolation of hardware traces, (2) cross-generator Medical Deepfake Detection, (3) scanner-invariant robustness, and (4) the visual explainability of the isolated K-Space Signatures.

\subsubsection{Pairwise Similarity Evaluation.}
To empirically validate that the extracted KSS tensors capture the traces of the involved devices, we perform a pairwise similarity analysis across all scanner models.

In detail, given two distinct scanner models, $A$ and $B$, their respective averaged three-dimensional tensors ($\tilde{P}_{A}$ and $\tilde{P}_{B}$) are flattened into one-dimensional vectors of length $N$ (the total number of voxels). We then evaluate their correlation using Cosine Similarity:

\begin{equation}
Sim(\tilde{P}_{A},\tilde{P}_{B}) = \frac{\sum_{i=1}^{N}\tilde{P}_{A,i}\cdot\tilde{P}_{B,i}}{\sqrt{\sum_{i=1}^{N}\tilde{P}_{A,i}^{2}} \cdot \sqrt{\sum_{i=1}^{N}\tilde{P}_{B,i}^{2}}}
\end{equation}

This metric effectively isolates the angular proximity of the high-dimensional noise signatures, measuring the correlation of spatial frequency patterns while ignoring global intensity scaling differences. Furthermore, as demonstrated in the Supplementary Materials, these intrinsic hardware topologies remain robustly preserved across spatial preprocessing workflows: cross-domain comparisons between native unaligned volumes and spatially registered counterparts confirm that resampling leaves the core frequency bands intact, while a comprehensive cross-interpolation ablation study further verifies that continuous spatial kernels securely maintain hardware-specific geometry against aggressive low-pass filtering.

\begin{figure}[htbp]
    \centering
    \includegraphics[width=0.95\linewidth]{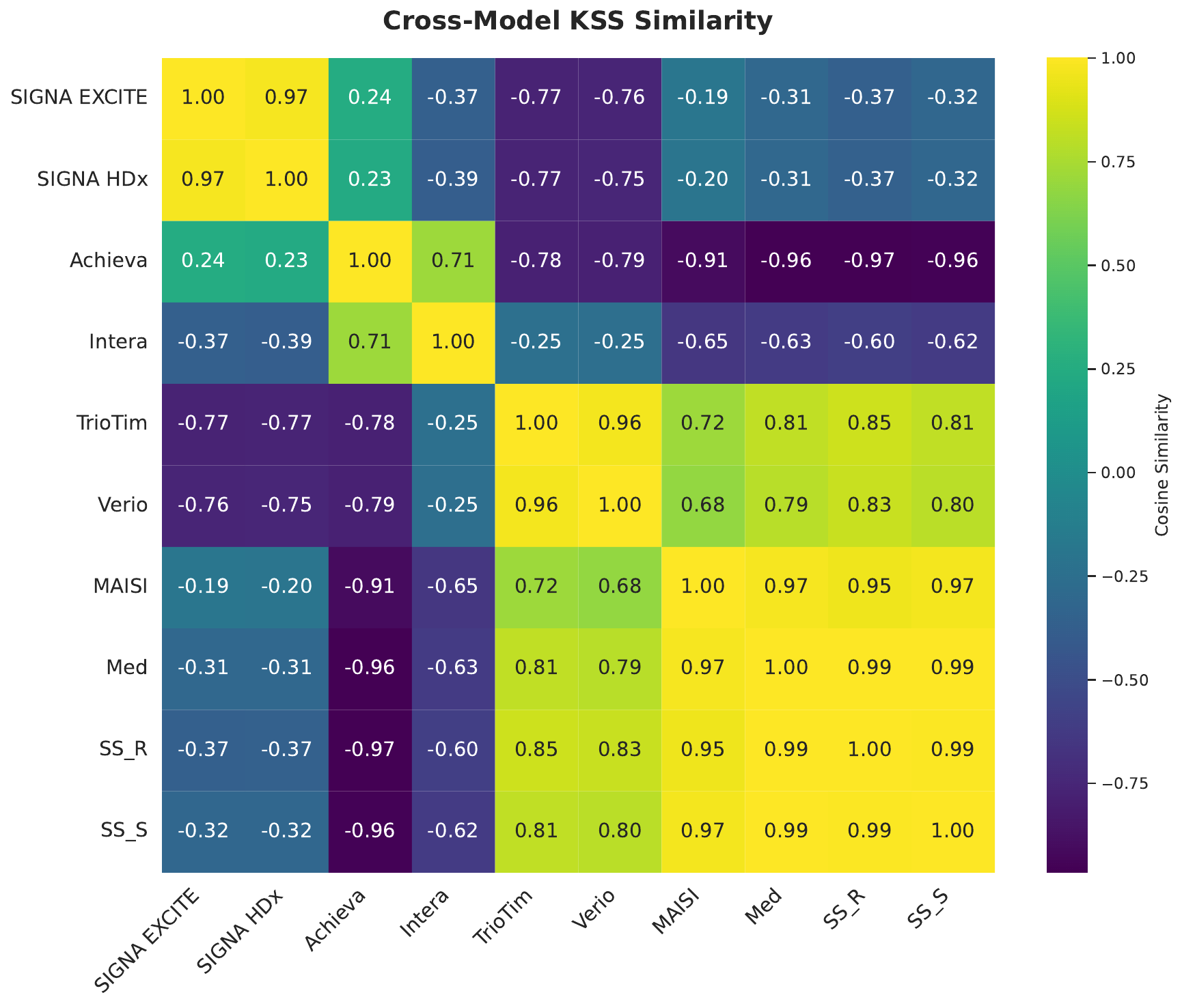}
    \caption{Cross-model cosine similarity matrix of isolated KSS (Log-PSD). Strong intra-manufacturer clusters (e.g., GE, Siemens) and inter-vendor negative correlations confirm highly distinct hardware domains. Notably, synthetic data strongly correlates with Siemens hardware, revealing an inherited scanner bias from the generative training process.}
    \label{fig:kss_similarity}
\end{figure}

\paragraph{Cross-Model Similarity and Hardware Affinity.}
The resulting cross-model similarity matrix, visualized in Fig.~\ref{fig:kss_similarity}, strongly validates our theoretical formulation. We observe highly distinct block-diagonal clusters indicating strong intra-manufacturer correlation. Specifically, GE scanners (SIGNA EXCITE and SIGNA HDx) share a near-identical noise topology with a similarity score of 0.97. A similarly robust clustering is evident for SIEMENS hardware (TrioTim and Verio) at 0.96, and to a notable extent for PHILIPS (Achieva and Intera) at 0.71.

\begin{figure*}[!ht]
    \centering
    \includegraphics[width=\textwidth]{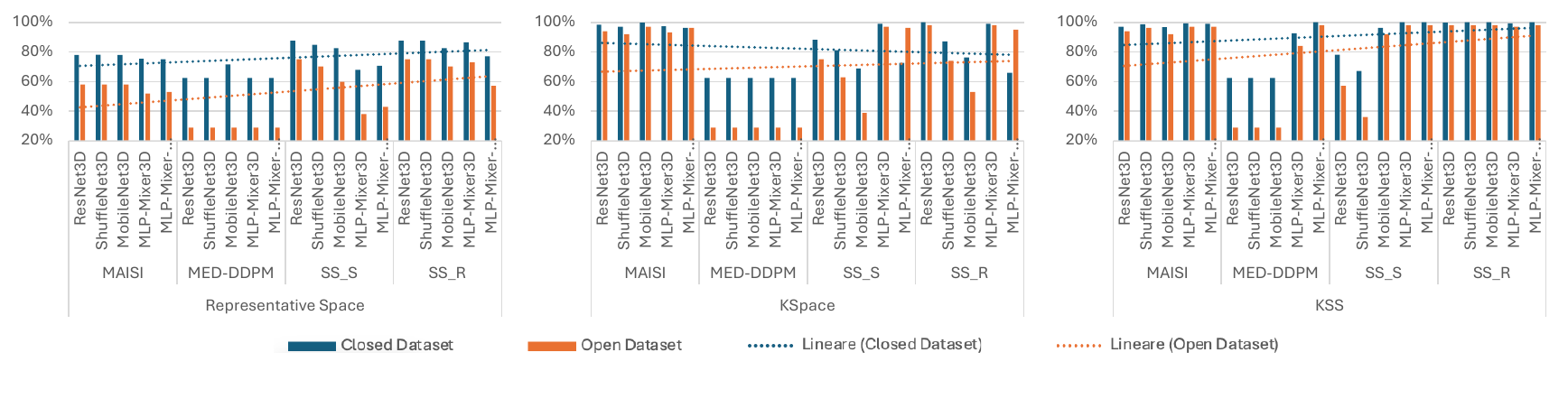} 
    \caption{Model accuracy comparison (Closed Dataset vs. Open Dataset) across architectures (ResNet3D, ShuffleNet3D, MobileNet3D, MLP-Mixer3D, MLP-Mixer-ArcFace3D). Evaluation spans three domains (Representative Space, K-Space, KSS) segmented by generative methods (MAISI, Med, SS\_S, SS\_R). Dashed lines show overall positive linear trends for Closed Dataset (blue) and Open Dataset (orange), which remain sensitive to the categorical x-axis ordering.}
    \label{res:resuls}
\end{figure*}

Crucially, the inter-vendor comparisons demonstrate significant orthogonality, definitively isolating the domains. For instance, the similarities between GE and SIEMENS collapse into strongly negative correlations (ranging from -0.75 to -0.77), while PHILIPS exhibits negative correlations with SIEMENS ranging from mild (-0.25 for Intera) to extreme (-0.79 for Achieva). This mathematical opposition is consistent with the hypothesis that different manufacturers inject fundamentally diverging noise distributions into the K-space, justifying our Leave-One-Domain-Out Multi-Source Domain Adaptation (MSDA) approach, which we explicitly evaluate in the subsequent Zero-Shot Domain Adaptation experiments. The complete MSDA framework, including the mathematical spatial reconstruction of KSS-filtered volumes via Inverse DFT across alternative training subsets, is further detailed in the Supplementary Materials. 

Furthermore, an analysis of the synthetic datasets (comprising MAISI, Med, and SuperSynth) reveals an insightful phenomenon. While displaying extreme intra-group similarity ($> 0.94$), the synthetic data exhibits a strong positive correlation specifically with SIEMENS hardware (0.67 to 0.84), alongside moderate to strong negative correlations with GE and PHILIPS. This suggests that the generative diffusion models utilized to synthesize these volumes inherently learned and replicated the hardware artifacts prevalent in their specific training distributions, embedding a permanent "Siemens-like" K-space bias into the generated synthetic anatomy.

\subsubsection{Medical Deepfake Detection.}
To evaluate the generalization capability and domain robustness of our framework, we conduct extensive cross-dataset experiments. The primary objective is to evaluate whether the proposed features allow standard and specialized neural network architectures to capture generative artifacts rather than memorizing domain-specific features. 
We formalize this task by partitioning the evaluation into three distinct representation spaces: raw spatial-domain MRIs, complex K-space magnitude representations, and our isolated KSS tensors. Let $\mathcal{M}$ represent the set of backbones evaluated, including 3D convolutional networks (ResNet3D, ShuffleNet3D, MobileNet3D) and Multi-Layer Perceptron networks (MLP-Mixer variants). For each domain $\mathcal{D} \in \{\text{MRI}, \text{K-Space}, \text{KSS}\}$, a model $f_m \in \mathcal{M}$ is trained via binary classification on a single generative dataset $\mathcal{S}_{\text{train}} \in \{\text{MAISI}, \text{Med}, \text{$SS_S$}, \text{$SS_R$}\}$ against the authentic training cohort. 
To test for generalization, each trained model is subjected to a rigorous Out-of-Distribution (OOD) evaluation against all alternative synthetic pipelines $\mathcal{S}_{\text{test}} \neq \mathcal{S}_{\text{train}}$ and a completely unseen Open Dataset containing real-world scans from novel scanner models.

\paragraph{Cross-Dataset Generalization Performance.}
The global accuracy trends across all architectural configurations and spatial domains are visualized in Fig.~\ref{res:resuls}, with the exhaustive cross-dataset evaluation metrics provided in the Supplementary Materials. While high classification performance on seen synthetic datasets indicates successful feature extraction, the key indicator of architectural robustness is localized within the Open Dataset evaluation. Models that maintain high accuracy on these unseen, real-world configurations demonstrate superior generalization, providing evidence that they represent the core structural discrepancies introduced by generators rather than overfitting to specific hardware noise patterns.
Our empirical findings reveal critical insights into the architectural inductive biases of the networks. Traditional 3D CNN backbones exhibit substantial vulnerability to domain shifts across generators, evidenced by severe accuracy drops (frequently falling below 35\%) on the Open Dataset when operating in the Representative Space and standard K-Space domains. Conversely, the MLP-Mixer variants, particularly when coupled with the KSS representation space, yield significantly higher performance stability. This global receptive field allows the network to process diffuse spectral anomalies distributed across the entire spectrum. 
As illustrated by the dense high-accuracy clusters in the KSS section of Fig.~\ref{res:resuls}, our frequency-domain backbone effectively minimizes the performance gap between the aggregate average (Closed Dataset) and the Open Dataset. Overall, these visual trends and the comprehensive supplementary metrics confirm the robustness of the KSS framework under challenging Out-of-Distribution (OOD) conditions and its viability for multi-scanner clinical deployment.

\subsubsection{Robustness and Scanner-Invariant Generalization.}
A primary challenge in medical image forensics is preventing the model from utilizing scanner-specific noise profiles (hardware artifacts) as a shortcut to distinguish between real and synthetic data. To rigorously evaluate the generalization capabilities of our best-performing KSS MLP-Mixer-ArcFace 3D architecture, we designed a strict leave-one-out domain generalization experiment.
In this formulation, the training distribution is intentionally constrained to a single clinical MRI manufacturer (e.g., exclusively GE, Philips, or Siemens) paired with only one synthetic dataset variant. The model is then evaluated on its ability to classify images derived from completely unseen generative models (Target Fakes) and entirely unseen clinical scanners (Open Set Scanners). This severe training constraint guarantees that the model must learn universal generative artifacts rather than overfitting to a specific hardware domain.

\paragraph{Scanner-Invariant Generalization Performance.}
Our results, detailed in Table~\ref{tab:robustness_generalization}, demonstrate robustness generalization capability. Despite being trained on a highly restricted distribution, the model consistently achieves nearly perfect PR-AUC and ROC-AUC scores (frequently approaching 1.000) when distinguishing all fakes against the source scanner. 
Crucially, while the AUC metrics confirm perfect class separability, a notable discrepancy in raw accuracy arises under specific extreme training conditions. For instance, when trained on the Philips scanner paired with the Med dataset, the model yields an accuracy of 55\% on Target Fakes despite maintaining 100\% PR-AUC and ROC-AUC. Cross-model KSS similarity analysis Fig.~\ref{fig:kss_similarity} reveals this to be a decision threshold artifact rather than a failure of discriminative power. The KSS of Med exhibits an extreme negative correlation with the Philips Achieva scanner ($r = -0.96$). Because these domains are virtually polar opposites, the network rapidly converges on a highly rigid, aggressive decision boundary during training. When evaluated on unseen Target Fakes, which are highly correlated with Med ($r \geq 0.95$) but retain slight distributional variances, the relative representation remains completely disjoint from real images (yielding perfect AUC). However, the absolute logit distributions shift just enough to cross the fixed binary decision threshold (e.g., 0.5), sharply reducing accuracy. This highlights the necessity of relying on threshold-independent metrics (ROC/PR-AUC) to accurately capture representation separability in severe domain generalization scenarios.
More importantly, the model maintains strong discriminative power (PR-AUC up to $93\%$) on the strictly unseen Open Set Scanners, though this drops to approximately $81\%$ when trained on the Med dataset. This confirms that the combination of Global Spectrum Subtraction (KSS) and ArcFace metric learning effectively forces the network to learn universal, scanner-invariant generative signatures rather than overfitting to specific MRI acquisition hardware, albeit with varying efficacy depending on the training dataset's domain shift.

\begin{table}[t]
    \centering
    \resizebox{\columnwidth}{!}{%
    \begin{tabular}{cl ccc ccc}
        \toprule
        \multirow{2}{*}{\rotatebox{90}{\textbf{Data}}} & \multirow{2}{*}{\textbf{Train Scanner}} & \multicolumn{3}{c}{\textbf{Target Fakes (All)}} & \multicolumn{3}{c}{\textbf{Open Set Scanners}} \\
        \cmidrule(lr){3-5} \cmidrule(lr){6-8}
        & & \textbf{Acc (\%)} & \textbf{PR-AUC (\%)} & \textbf{ROC-AUC (\%)} & \textbf{Acc (\%)} & \textbf{PR-AUC (\%)} & \textbf{ROC-AUC (\%)} \\
        \midrule
        \multirow{3}{*}{\rotatebox{90}{\textbf{MAISI}}} 
        & GE      & 100 & 100 & 100 & 88 & 93 & 79 \\
        & PHILIPS & 100 & 100 & 100 & 89 & 93 & 79 \\
        & SIEMENS & 99  & 100 & 100 & 89 & 93 & 79 \\
        \midrule
        \multirow{3}{*}{\rotatebox{90}{\textbf{Med}}} 
        & GE      & 100 & 100 & 100 & 93 & 81 & 69 \\
        & PHILIPS & 55  & 100 & 100 & 49 & 81 & 68 \\
        & SIEMENS & 99  & 100 & 100 & 89 & 82 & 69 \\
        \midrule
        \multirow{3}{*}{\rotatebox{90}{\textbf{SS\_R}}} 
        & GE      & 92  & 100 & 100 & 88 & 93 & 79 \\
        & PHILIPS & 100 & 100 & 100 & 93 & 93 & 79 \\
        & SIEMENS & 98  & 100 & 100 & 88 & 93 & 79 \\
        \midrule
        \multirow{3}{*}{\rotatebox{90}{\textbf{SS\_S}}} 
        & GE      & 99  & 100 & 100 & 88 & 93 & 79 \\
        & PHILIPS & 82  & 100 & 100 & 76 & 93 & 79 \\
        & SIEMENS & 99  & 100 & 100 & 88 & 93 & 79 \\
        \bottomrule
    \end{tabular}
    }
    \caption{Generalization and robustness evaluation. During training, the KSS MLP-Mixer-ArcFace model was restricted to a single MRI manufacturer and a single generative model dataset. Performance is evaluated across all available synthetic datasets, excluding the training set (Target Fakes), and on completely unseen clinical devices (Open Set Scanners).}
    \label{tab:robustness_generalization}
\end{table}

\subsection{Visual Explainability of the KSS}
To demonstrate that KSS disentangles hardware traces from patient anatomy, we introduce a reverse-mapping protocol. By dynamically modulating the isolated KSS within the logarithmic frequency domain and projecting it back into the spatial representation, we can visually amplify or suppress the hardware trace.

\begin{figure}[t]
    \centering
    \includegraphics[width=\linewidth]{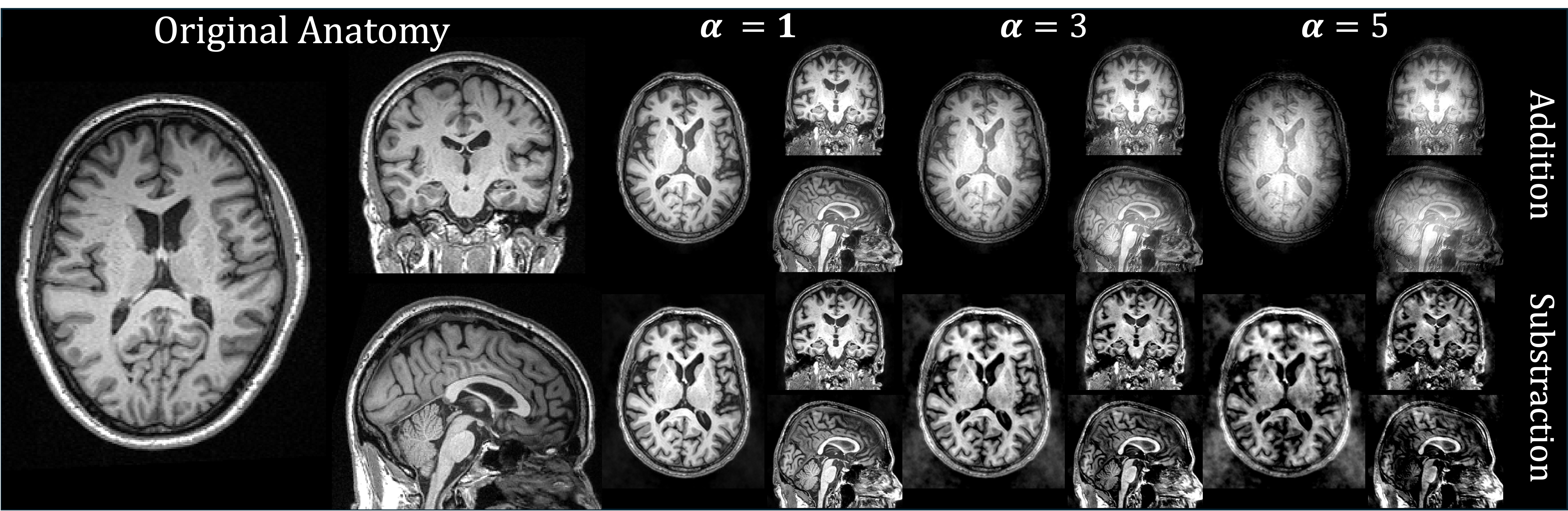} 
    \caption{Triplanar reconstructions demonstrating that macroscopic patient anatomy remains structurally intact during KSS isolation. By applying a scalar multiplier $\alpha$ in the logarithmic frequency domain $(\tilde{P}_{mod}=\tilde{P}_{orig}\pm\alpha\cdot KSS)$, we amplify or suppress hardware-specific spectral footprints to visually expose generative traces across manipulation conditions $\alpha\in\{1,3,5\}$.}
    \label{fig:anatomy_preservation}
\end{figure}

Crucially, as illustrated by the triplanar spatial reconstructions in Fig.~\ref{fig:anatomy_preservation}, the macroscopic anatomical structures (e.g., ventricles, sulci, and cortical boundaries) remain structurally intact under these spectral manipulations. This visual preservation proves that the network does not rely on spatial anatomical shortcuts. The complete mathematical formulation of this reverse-mapping, alongside comprehensive visual evaluations across extreme manipulation conditions and all scanner typologies, is provided in the Supplementary Materials.

\section{Discussion and Limitations}
This work introduces the K-Space Signature (KSS), a novel frequency-domain framework designed to detect fully synthetic Medical Deepfakes and protect clinical diagnostic pipelines. By shifting the forensic analysis to the Logarithmic Power Spectral Density (Log-PSD) space and explicitly subtracting an empirical global anatomical baseline ($\tilde{P}_{global}$), the proposed method successfully isolates intrinsic hardware acquisition noise and generative algorithmic traces from macroscopic patient anatomy. To process these globally distributed spectral signatures without local spatial bias, the framework utilizes a specialized 3D MLP-Mixer architecture equipped with an ArcFace spherical metric-learning head. Extensive cross-dataset evaluations confirm that the KSS framework provides a highly effective and robust defense mechanism. The approach demonstrates exceptional detection performance, exceeding 0.99 Accuracy and ROC-AUC across multi-generator synthetic datasets. Furthermore, it exhibits robust zero-shot domain generalization, maintaining strong discriminative power (up to 0.93 Accuracy) when evaluated on independent datasets from completely unseen clinical scanners and alternative generative models.While KSS successfully captures macroscopic spectral deviations, generative pipelines and clinical hardware leave multifaceted footprints that warrant further exploration. The primary limitation of this study is the constrained dataset size; finding large-scale, public medical datasets with comprehensive scanner metadata remains a significant challenge for robust evaluation. Additionally, the current formulation requires a representative cohort of authentic training volumes to compute the $\tilde{P}_{global}$ prior before isolation.  Future research will focus on eliminating the reliance on a pre-computed anatomical prior by developing dynamic, single-volume feature extraction techniques. We also aim to expand the framework's diagnostic scope to encompass varying MRI contrast weightings and other 3D modalities, such as PET and CT. Finally, we intend to transition from global binary classification to targeted spectral analysis, enabling the precise mapping of synthetic frequencies artificially implanted into forgeries. Ultimately, by mathematically disentangling physical and generative artifacts in the spectral domain, this research provides a comprehensive foundation for medical data authenticity, capable of both neutralizing synthetic forgeries and reliably attributing authentic scans to their precise hardware origins.
\bibliography{references}

\clearpage

\appendix
\section*{Supplementary Material}
To validate the proposed framework, this supplementary material details the experimental setup and evaluates the theoretical robustness, spectral integrity, and generalization capabilities of the K-Space Signature (KSS) isolation strategy. Specifically, the extended evaluation is partitioned into the following core sections:

\begin{itemize}
    \item \textbf{A. Datasets}
    \begin{itemize}
        \item Details the presented real MRI cohort (ADNI~\cite{ADNI_2010} and PPMI~\cite{PPMI_2011}), the synthetic dataset (generated via MAISI-v2~\cite{maisi}, SuperSynth~\cite{billot2023synthseg}, and Med-DDPM~\cite{med-ddpm}), and the open-set dataset designed exclusively for zero-shot testing to evaluate generalization.
\end{itemize}
    
    \item \textbf{B. Implementation Details}
    \begin{itemize}
        \item Provides the hardware specifications and the exact training hyperparameters applied across all network architectures.
    \end{itemize}
    
    \item \textbf{C. Ablation Study}
    \begin{itemize}
    \item \textbf{Empirical Robustness:} Validates using the global mean ($\tilde{P}_{global}$) to efficiently remove shared anatomical structures.
    \item \textbf{Spectral Integrity \& Resampling:} Confirms that spatial registration and continuous interpolation kernels (e.g., Lanczos) preserve unique scanner noise topologies.
    \item \textbf{Extended Performance:} Provides granular metrics demonstrating the stability of MLP-Mixer-ArcFace3D and the KSS space across 60 independent configurations.
    \item \textbf{Zero-Shot Adaptation:} Evaluates generalization capabilities across unseen scanner manufacturers.
    \item \textbf{Visual Explainability:} Uses a scalar multiplier ($\alpha$) to visually demonstrate feature disentanglement without anatomical distortion.
\end{itemize}
\end{itemize}

\section{Datasets}
To rigorously prevent data leakage and ensure an unbiased evaluation of hardware traces, we partition the dataset into training, validation, and testing splits (80:10:10) strictly at the patient level. Specifically, all MRI volumes belonging to a single patient, even if acquired across different scanner models, are constrained to a single subset. This subject-wise isolation minimizes the risk that the model uses patient-specific anatomical memorization as a shortcut during evaluation. The splits are further stratified to maintain a balanced distribution of authentic MRI scans across all subsets. Furthermore, we enforce strict isolation protocols for the synthetic data. While unconditionally generated synthetic volumes are partitioned randomly, synthetic volumes produced via Image-to-Image (I2I) translation pipelines are deterministically assigned to the exact same split as their authentic source volume. This ensures that data leakage is strictly controlled to prevent inadvertent exposure to the model during training through a synthetically altered counterpart.

\subsection{Real Dataset}

The real dataset comprises 1,200 T1-weighted brain MRI scans collected from the Image and Data Archive (IDA) at the Laboratory of Neuro Imaging (LONI), including 1,128 scans from the ADNI and 72 scans from the PPMI. These scans were randomly subsampled from the larger database to achieve a perfectly balanced distribution across acquisition hardware. Overall, the dataset contains 687 unique subjects (372 males and 315 females). The average age at the time of the first scan is $73.44 \pm 7.59$ years. Specifically, the male population has an average age of $74.71 \pm 7.49$ years, while the female population averages $71.93 \pm 7.45$ years. 

The population distribution by Research Group for the unique subjects is as follows: Cognitively Normal (CN): 160, Mild Cognitive Impairment (MCI): 159, Early MCI (EMCI): 130, Alzheimer's Disease (AD): 91, Late MCI (LMCI): 54, Prodromal: 53, Significant Memory Concern (SMC): 26, Parkinson's Disease (PD): 12, and Control: 2. In terms of total scan counts, the distribution spans: CN: 303, MCI: 297, EMCI: 243, AD: 141, LMCI: 100, Prodromal: 58, SMC: 44, PD: 12, and Control: 2.

To rigorously evaluate the harmonization pipeline while minimizing potential confounding factors, the dataset was carefully balanced across three major MRI manufacturers, allocating 400 scans to each vendor: GE MEDICAL SYSTEMS, Philips Medical Systems, and SIEMENS. Each manufacturer is equally represented by two scanner models (200 scans each):
\begin{itemize}
    \item GE MEDICAL SYSTEMS: SIGNA EXCITE (1.5T / 3.0T) and SIGNA HDx (1.5T / 3.0T).
    \item Philips Medical Systems: Achieva (1.5T / 3.0T) and Intera (1.5T).
    \item SIEMENS: TrioTim (3.0T) and Verio (3.0T).
\end{itemize}
This balanced design mitigates the risk of scanner-specific traces being confounded by class imbalance, allowing the proposed framework to accurately characterize intrinsic hardware-specific traces.

\subsection{Synthetic Dataset}
To evaluate the robustness of the proposed KSS under diverse generation paradigms, we constructed a synthetic dataset using three state-of-the-art 3D MRI synthesis frameworks: MAISI-v2~\cite{maisi} (MAISI), SuperSynth~\cite{billot2023synthseg} (providing reampled geometry, SS\_R, and synthetic contrast, SS\_S) specifically version 2025, and Med-DDPM~\cite{med-ddpm} (Med). This selection intentionally encompasses unconditional generation (MAISI), image-to-image contrast translation (SS\_R, SS\_S), and conditioned diffusion (Med).

MAISI-v2 was employed to generate 1,200 fully synthetic MRI volumes. SuperSynth was applied to the real cohort to extract anatomical segmentation masks and produce paired synthetic contrast-standardized volumes. Finally, the extracted segmentation masks were used as structural priors for Med-DDPM, generating an additional set of 1,200 synthetic MRI volumes that preserve the anatomical geometry of the original subjects while introducing entirely synthetic imaging characteristics. This multi-generator cohort provides a rigorous benchmark to evaluate whether our framework captures universal, generator-agnostic spectral traces rather than overfitting to a specific synthesis pipeline.

\subsection{Open-Set Evaluation Dataset}

To assess the zero-shot generalization capability of the proposed framework, we curated an independent \textit{Open-Set Evaluation Dataset} comprising 120 authentic MRI scans acquired from scanner models never observed during training or validation. The dataset is uniformly balanced across the same three manufacturers, including 40 scans from each previously unseen scanner model: GENESIS SIGNA (GE,1.5T), MR 7700 (Philips,3.0T) and Symphony (Siemens,1.5T).

This dataset is used exclusively during inference to evaluate the ability of the proposed framework to generalize across unseen acquisition hardware and verify that the learned representations capture generator-related traces rather than memorizing scanner-specific traces.

Finally, all MRI volumes were aligned to the MNI152 T1 1mm utilizing ANTsPy (version 0.6.3)~\cite{avants2011reproducible} for the \texttt{antsRegistration} routine with Lanczos interpolation. Rigid registration was strictly utilized to align the volumes to the standard template. This deliberate constraint, restricting transformations to translation and rotation while avoiding scaling or affine warping, was implemented to preserve the native high-frequency spectral geometry. By preventing scale-induced frequency shifts and minimizing destructive interpolation filtering, this approach facilitates the extraction of a KSS that reflects intrinsic hardware and generative artifacts rather than differential preprocessing distortions. While spatial interpolation inherently acts as a frequency filter, this preprocessing step is uniformly applied across all authentic and synthetic volumes. Consequently, any systematic frequency shift introduced by the Lanczos kernel remains strictly constant across the dataset, allowing the isolated features to reliably reflect the underlying traces. A comprehensive empirical analysis of this interpolation effect is detailed in the subsequent ablation studies.  

Additionally, as explained in the main paper, due to the limited number of generative models, the networks were trained using only a single synthetic dataset at a time. To evaluate cross-generator generalization, all remaining synthetic datasets excluded from the training phase were strictly utilized as unseen, open-set testing domains.

\section{Implementation Details}
\subsection{Hardware}
All experiments were conducted on a dedicated workstation equipped with an NVIDIA A100 Tensor Core GPU and an Intel Xeon E5-2650 v4 CPU.

\subsection{Training Hyperparameters for GLOBAL ablation study}
To evaluate the classification performance across the 23 alternative baseline formulations reported in Table~\ref{tab:model_results}, we utilized a standard 3D MLP-Mixer architecture. For this specific ablation, the ArcFace metric-learning head was intentionally removed, and a standard linear classification head with Cross-Entropy Loss was employed. This ablation design isolates the intrinsic structural integrity of the extracted KSS representations, ensuring that the results reflect the validity of the baseline subtraction rather than the separating power of complex angular margin constraints. 

The network hyperparameters were strictly fixed across all baseline evaluations:
\begin{itemize}
    \item \textbf{Optimization:} AdamW optimizer with a learning rate of $5 \times 10^{-5}$ and a CosineAnnealingLR scheduler.
    \item \textbf{Loss Function:} CrossEntropyLoss with a label smoothing of $0.1$.
    \item \textbf{Architecture:} Patch size of $16$, hidden dimension (\texttt{dim}) of $128$, depth of $6$, token MLP dimension of $256$, channel MLP dimension of $512$, and a dropout rate of $0.4$.
    \item \textbf{Training Dynamics:} Batch size of $64$ for a maximum of $100$ epochs. Early stopping was implemented with a patience count of $10$ and a minimum delta of $0.001$. Empirically, all ablation models converged and triggered early stopping around the $20^{\text{th}}$ epoch.
\end{itemize}

\subsection{Training Hyperparameters for Cross-Dataset Evaluation}
To generate the comprehensive cross-dataset evaluation results reported in Table~\ref{tab:cross_dataset_results_split}, we established standardized training protocols for both the baseline 3D Convolutional Neural Networks (MobileNet3D, ShuffleNet3D, ResNet3D) and the proposed MLP-Mixer3D variants. 

Across all architectures, the optimization strategy remained consistent. We utilized the AdamW optimizer with a weight decay of $0.05$, paired with a CosineAnnealingLR scheduler ($T_{max} = 500$, $\eta_{min} = 1 \times 10^{-6}$). The training loops were constrained to a maximum of $500$ epochs, incorporating an early stopping mechanism with a patience of $10$ epochs and a minimum delta of $0.001$. Label smoothing was uniformly applied at $0.1$.

\subsubsection{MLP-Mixer3D Architectures}
For the MLP-Mixer3D and MLP-Mixer-arcface3D models, training was conducted using a batch size of $64$ and a learning rate of $0.0001$. The core architectural hyperparameters were defined as follows: patch size of $16$, embedding dimension (\texttt{dim}) of $128$, depth of $6$, token MLP dimension of $256$, channel MLP dimension of $512$, and a dropout rate of $0.4$. Because these networks were configured with a single output class dimension, \texttt{BCEWithLogitsLoss}~\cite{paszke2019pytorch} was utilized as the classification criterion. 

For the ArcFace variant specifically, the spherical metric head was parameterized with a scaling factor $s = 30.0$ and an angular margin $m = 0.5$.

\subsubsection{Baseline 3D CNN Architectures}
For the baseline convolutional networks (MobileNet3D, ShuffleNet3D, and ResNet3D), the hyperparameter configuration was adjusted to accommodate the architectural differences. The training batch size was set to $6$, with a learning rate of $5 \times 10^{-5}$. Unlike the MLP-Mixer configurations, these models were set to output $2$ classes, and consequently, \texttt{CrossEntropyLoss}~\cite{paszke2019pytorch} was utilized as the classification criterion.

\begin{table}[t!]
    \centering
    \resizebox{\columnwidth}{!}{%
    \begin{tabular}{lcccc}
        \toprule
        \textbf{Model} & \textbf{Accuracy (\%)} & \textbf{ROC AUC (\%)} & \textbf{F1 (\%)} & \textbf{Precision (\%)} \\
        \midrule
        Scanner 0 & 71.7 (+5.0) & 94.9 (+0.2) & 66.1 (-0.1) & 63.5 (-3.4) \\
        Scanner 1 & 65.0 (-1.7) & \textit{93.9} (-0.8) & 63.8 (-2.4) & 66.9 (0.0) \\
        Scanner 2 & 69.4 (+2.7) & 95.2 (+0.5) & 68.6 (+2.4) & 70.4 (+3.5) \\
        Scanner 3 & 68.3 (+1.6) & 95.0 (+0.3) & 64.5 (-1.7) & 68.2 (+1.3) \\
        Scanner 4 & 68.9 (+2.2) & 93.7 (-1.0) & 67.9 (+1.7) & 70.7 (+3.8) \\
        Scanner 5 & \textit{64.4} (-2.3) & 94.0 (-0.7) & \textit{59.2} (-7.0) & \textit{56.3} (-10.6) \\
        COMB 0 1  & 71.7 (+5.0) & 95.1 (+0.4) & 70.4 (+4.2) & 73.3 (+6.4) \\
        COMB 0 2  & 69.4 (+2.7) & 96.0 (+1.3) & 69.3 (+3.1) & 69.9 (+3.0) \\
        COMB 0 3  & 71.1 (+4.4) & 94.3 (-0.4) & 70.1 (+3.9) & 71.5 (+4.6) \\
        COMB 0 4  & 72.2 (+5.5) & 94.4 (-0.3) & 71.3 (+5.1) & 74.3 (+7.4) \\
        COMB 0 5  & \textbf{74.4 (+7.7)} & 94.5 (-0.2) & 72.2 (+6.0) & 79.8 (+12.9) \\
        COMB 1 2  & 65.0 (-1.7) & 93.9 (-0.8) & 64.0 (-2.2) & 64.5 (-2.4) \\
        COMB 1 3  & 72.8 (+6.1) & 95.0 (+0.3) & 72.5 (+6.3) & 73.4 (+6.5) \\
        COMB 1 4  & 73.3 (+6.6) & 95.3 (+0.6) & \textbf{73.2} (+7.0) & 73.8 (+6.9) \\
        COMB 1 5  & 71.1 (+4.4) & 95.4 (+0.7) & 69.8 (+3.6) & 72.9 (+6.0) \\
        COMB 2 3  & 70.6 (+3.9) & 94.9 (+0.2) & 70.3 (+4.1) & 71.1 (+4.2) \\
        COMB 2 4  & 72.8 (+6.1) & \textbf{95.8} (+1.1) & 72.3 (+6.1) & 74.6 (+7.7) \\
        COMB 2 5  & 68.9 (+2.2) & 94.5 (-0.2) & 68.2 (+2.0) & 68.8 (+1.9) \\
        COMB 3 4  & 69.4 (+2.7) & 94.8 (+0.1) & 67.9 (+1.7) & 71.4 (+4.5) \\
        COMB 3 5  & 70.0 (+3.3) & 94.7 (0.0) & 68.8 (+2.6) & 71.5 (+4.6) \\
        COMB 4 5  & 70.0 (+3.3) & 94.2 (-0.5) & 66.6 (+0.4) & \textbf{76.8} (+9.9) \\
        SPLIT 1   & 69.4 (+2.7) & 95.1 (+0.4) & 67.3 (+1.1) & 69.8 (+2.9) \\
        SPLIT 2   & 72.8 (+6.1) & 95.3 (+0.6) & 71.1 (+4.9) & 77.2 (+10.3) \\
        GLOBAL    & 66.7 & 94.7 & 66.2 & 66.9 \\
        \midrule
        \textbf{Average} & 70.1 & 94.8 & 68.5 & 70.9 \\
        \textbf{Std Dev ($\sigma)$} & 2.7 & 0.6 & 3.4 & 5.0 \\
        \bottomrule
    \end{tabular}
    }
     \caption{Performance Comparison of KSS Extraction Baselines. Deltas ($\Delta$), reported above all percentage values, indicate the difference relative to the proposed GLOBAL baseline configuration.}
    \label{tab:model_results}
\end{table}

\section{Ablation Study}
\subsection{Empirical Robustness of the Global Baseline}

A potential theoretical concern regarding our KSS isolation strategy is whether subtracting a singular global mean ($\tilde{P}_{global}$) introduces systemic biases or dilutes scanner-specific hardware traces, as opposed to utilizing more granular, sub-population baselines. To rigorously validate our choice of a global baseline, we conduct an ablation study evaluating the model's classification performance across 23 alternative baseline formulations. 

Specifically, instead of a single universal average, we compute the anatomical subtraction term by isolating individual scanners (Scanners 0--5), various multi-scanner combinations (COMB), and distinct data splits (SPLIT) consisting of a single scanner from each manufacturer. We then extract the KSS utilizing these alternative baselines and evaluate the resulting impact on deepfake detection performance.

\subsubsection{Results and Variance Analysis}
As reported in Table~\ref{tab:model_results}, the variation in predictive performance across all 23 baseline configurations, which include individual isolated scanners (Scanners 0-5), multi-scanner combinations (COMB), and distinct data splits (SPLIT), is remarkably minimal. To provide a comprehensive evaluation, the table details several key internal variables: Accuracy (\%), ROC AUC (\%), F1 (\%), and Precision (\%), alongside their respective deltas ($\Delta$) indicating the difference relative to the proposed GLOBAL baseline configuration. For clarity in interpreting these internal variables, the best-performing values across the metrics are highlighted in bold, while the minor or lowest values are denoted in italics.  The standard deviations across the key performance metrics are near zero, yielding $\sigma=0.006$ for ROC AUC and $\sigma=0.027$ for Accuracy. This stability is consistent across the board, anchored by an average Accuracy of 70.1\% and a high average ROC AUC of 94.8\%. Furthermore, the performance of the proposed GLOBAL subtraction remains tightly clustered with the highly specialized configurations, with metric $\Delta$ typically bounded within marginal fractions.  This extremely low variance empirically validates our mathematical formulation for KSS isolation. It demonstrates that the KSS extraction is highly robust: the global average successfully suppresses the shared anatomical structural information without introducing configuration-dependent artifacts. Consequently, this suggests that $\tilde{P}_{global}$ is both a computationally efficient and structurally sound approximation of the universal human neuroanatomical baseline, requiring no dataset-specific tuning. 

Additionally, it is pertinent to note that this ablation configuration inherently performs a 6-class scanner classification task. When the predicted classes are aggregated at the broader manufacturer level, the predictive performance approaches $99\%$, consistent with state-of-the-art models within the research community. As a concise conclusion, these findings indicate that while the isolated KSS does not retain sufficient granular information for perfect intra-vendor scanner classification, it robustly preserves the structural spectral footprint necessary for highly accurate manufacturer identification.

\subsection{Registration Comparison and Spectral Integrity}

While the proposed framework successfully isolates hardware and generative traces, a theoretical concern regarding our pipeline is the extraction of the KSS subsequent to spatial registration. Although Lanczos interpolation is applied uniformly across all datasets to ensure an unbiased comparative evaluation, spatial resampling inherently acts as a complex frequency filter. To rigorously demonstrate that this preprocessing step does not dilute the intrinsic hardware traces, we formalize a cross-domain empirical analysis between native (RAW) and spatially registered (REG) MRI volumes.

Specifically, for each physical scanner model $m$, we extract the scanner-associated spectral trace from the unregistered native-space volumes, denoted as $\text{KSS}_{\text{raw}, m}$, and compare it directly against its spatially registered counterpart, $\text{KSS}_{\text{reg}, m}$. To ensure strict dimensional consistency for metric evaluation without altering the zero-frequency center, a deterministic central spatial crop is applied to align the volumetric tensors prior to frequency mapping. We evaluate the structural preservation of the traces by computing the pairwise Cosine Similarity between any spatially registered trace $i$ and any native-space trace $j$. To explicitly quantify the localized frequency deviations, we define the Log-PSD residual for any given cross-domain pair as:

\begin{equation}
    \hat{\Delta}_{i, j} = \text{KSS}_{\text{reg}, i} - \text{KSS}_{\text{raw}, j}
    \label{eq:regvsraw}
\end{equation}

To isolate the pure spectral footprint of the spatial interpolation without confounding hardware variations, we restrict our residual analysis to the intra-scanner cases where $i = j$.
\subsubsection{Baseline Hardware traces (RAW Domain)}
\begin{figure}[t!]
    \centering
    \includegraphics[width=0.95\columnwidth]{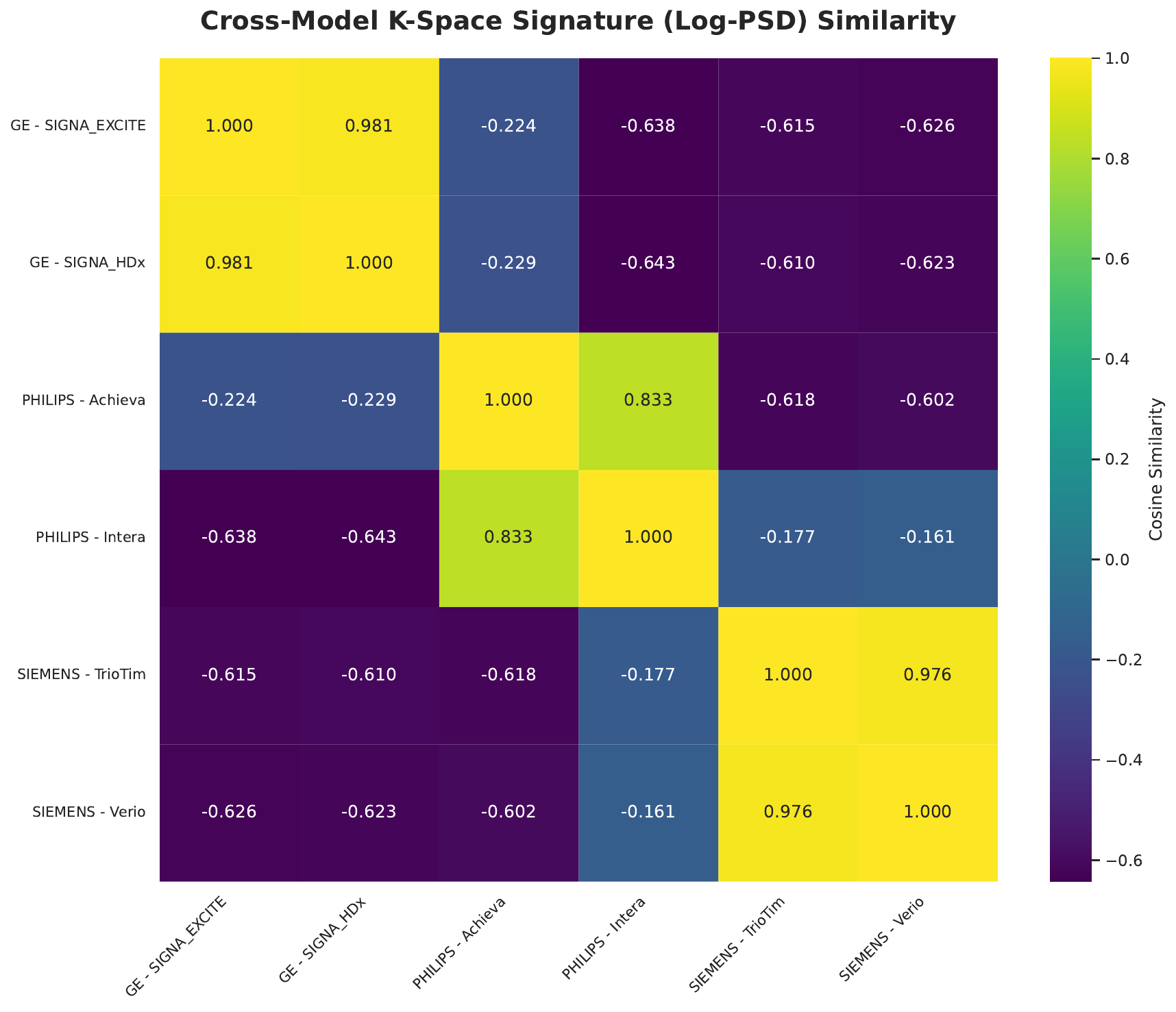}
    \caption{Cross-model Cosine Similarity matrix of the isolated KSSs computed strictly on native spatial configurations (RAW). The matrix confirms that intra-manufacturer harmony and inter-manufacturer orthogonality exist inherently before any spatial preprocessing.}
    \label{fig:reg_raw_heatmap}
\end{figure}

We first establish that the discriminative hardware traces exist inherently within the native acquisition space, independent of any spatial alignment. We compute the cross-model similarity matrix strictly on the RAW unaligned volumes for all six clinical scanner models (Fig.~\ref{fig:reg_raw_heatmap}). 
The empirical results reveal that the core topological properties of the hardware domains are highly pronounced in the native frequency space. We observe strong intra-manufacturer block-diagonal clusters (e.g., GE SIGNA models yield a similarity of $0.981$, and Siemens models yield $0.976$). Simultaneously, the inter-manufacturer comparisons display severe orthogonality, with correlations ranging between $-0.60$ and $-0.85$. This confirms that distinct physical noise topologies are intrinsically embedded during acquisition.

\begin{figure}[t!]
    \centering
    \includegraphics[width=0.95\columnwidth]{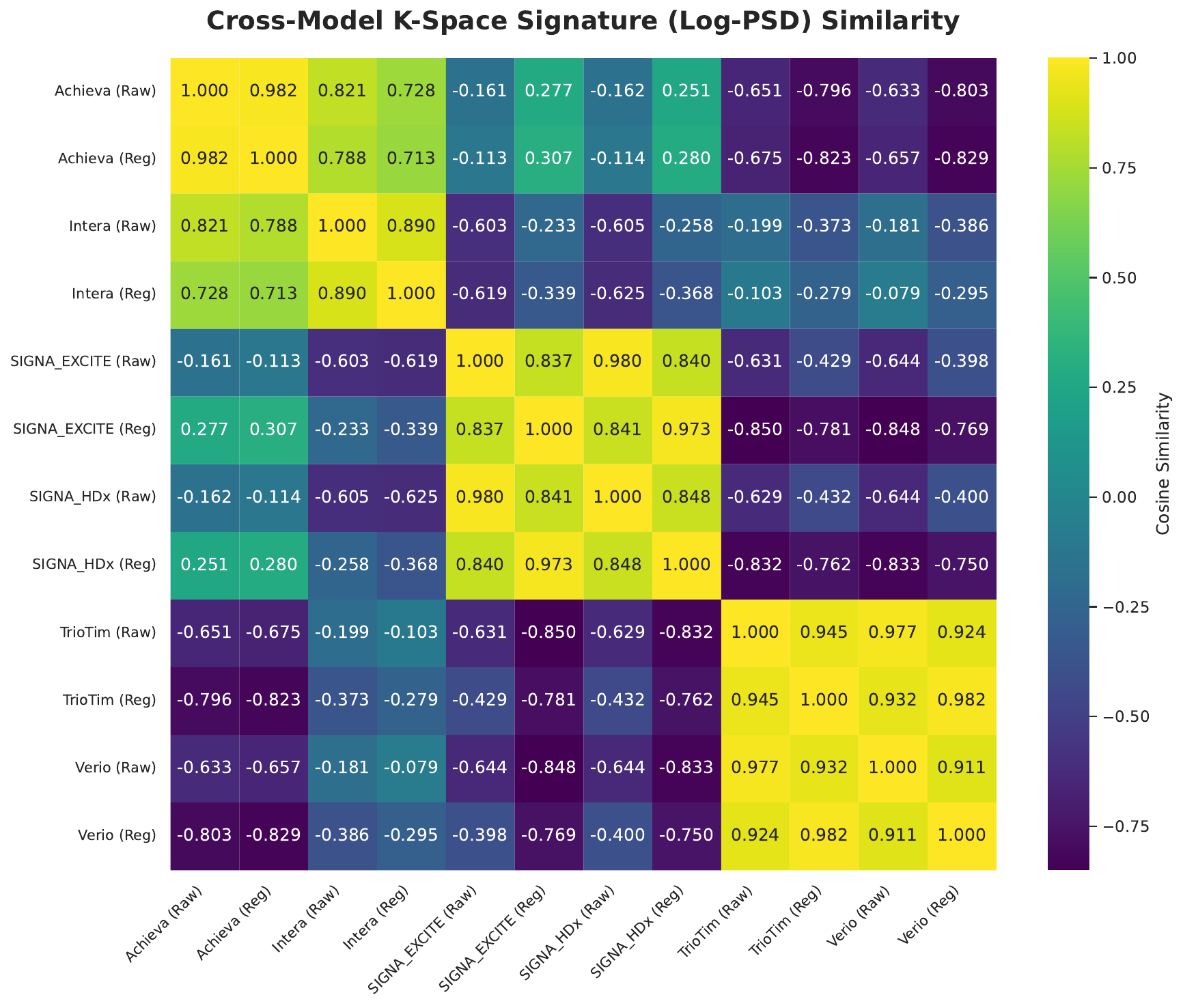}
    \caption{Extended cross-domain similarity matrix comparing RAW and REG KSS tensors. The high similarity scores along the principal diagonal block pairings demonstrate that spatial registration preserves the unique noise topology of each specific scanner model.}
    \label{fig:reg_comparison_heatmap}
\end{figure}

\subsubsection{Cross-Domain Alignment (RAW vs. REG)}
To observe how these intrinsic traces interact with the preprocessing pipeline, we compute an extended $12 \times 12$ cross-domain similarity matrix (Eq.~\ref{eq:regvsraw}) incorporating both RAW and REG representations (Fig.~\ref{fig:reg_comparison_heatmap}). The results yield exceptional intra-scanner coherence across domains. The direct correlation between the RAW and REG versions of the exact same physical scanner reaches $0.982$ for the Philips Achieva, $0.980$ for the GE SIGNA EXCITE, and $0.945$ for the Siemens TrioTim. Furthermore, the structurally antagonistic relationship between different manufacturers is perfectly retained across domains (e.g., Siemens TrioTim REG maintains a strong negative correlation of $-0.796$ against Philips Achieva RAW). This invariant orthogonality strongly indicates that the spatial transformation does not overwrite or scramble the underlying noise topology, allowing the downstream network to reliably extract hardware-invariant algorithmic traces.

\subsubsection{Quantifying the Interpolation Filter Trace}
To mathematically explain why the spatial resampling does not degrade the KSS discriminative power, we isolate the exact spectral impact of the Lanczos interpolation kernel. We explicitly define the spectral footprint of the registration pipeline as the volumetric difference tensor, $\Delta_{\text{reg}}$:

\begin{equation}
\Delta_{\text{reg}} = \text{KSS}_{\text{reg}} - \text{KSS}_{\text{raw}}
\end{equation}

We then track the 1D radial profile of this tensor averaged across the entire clinical cohort. 

\begin{figure}[t!]
    \centering
    \includegraphics[width=0.95\columnwidth]{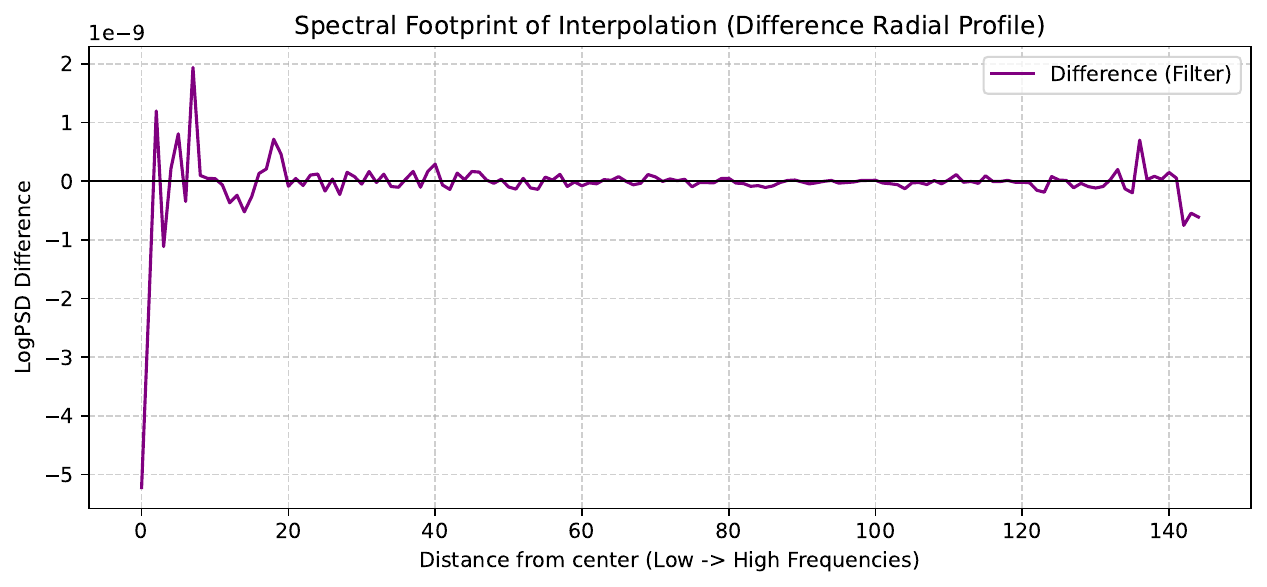}
    \caption{Spectral footprint of the registration pipeline. The 1D radial profile of the difference tensor ($\Delta_{\text{reg}}$) demonstrates that the registration filter leaves the core frequency bands highly preserved, with deviations restricted solely to the extreme high-frequency peripheral tails.}
    \label{fig:reg_radial_diff}
\end{figure}

As visualized in Fig.~\ref{fig:reg_radial_diff}, the spatial interpolation acts strictly as a localized high-frequency attenuation pass-band filter. The residual energy of the difference remains virtually zero throughout the vast majority of the low- and mid-frequency spectrum (spanning from the DC center up to a radial distance of approximately $100$ voxels). A controlled spectral deviation occurs exclusively at the absolute high-frequency tail of the spectrum (radial distances $> 120$ voxels), which represents the smoothing of volatile, pixel-level microscopic noise. 

Because the structural disparities introduced by physical scanners and synthetic generative pipelines are distributed globally across the entire spectrum, they remain structurally intact. Consequently, this demonstrates that while spatial registration introduces a localized high-frequency suppression, it does not alter the fundamental spectral geometry required for robust zero-shot medical deepfake detection.

\subsection{Impact of Spatial Resampling Typologies}

A critical concern in frequency-domain analysis is whether the network inadvertently leverages artifacts introduced by spatial interpolation as a shortcut to resolve the classification task. To prevent this, our alignment pipeline strictly utilizes rigid transformations, restricting operations to translation and rotation while explicitly avoiding scaling or affine warping that could distort the native high-frequency geometry. Furthermore, while our primary pipeline employs a Lanczos kernel to preserve high-frequency details, standard clinical workflows frequently rely on alternative interpolation techniques. To systematically prove that the discriminative power of the KSS stems from intrinsic hardware traces rather than the specific footprint of the applied filter, we conduct a comprehensive cross-interpolation ablation study.

We expand our preprocessing pipeline to include a set of four distinct interpolation kernels: $T$ = \{\text{Lanczos}, \text{Linear}, \text{Nearest Neighbor}, \text{B-Spline}\}. We evaluate the framework across two primary dimensions:

\begin{itemize}
    \item \textbf{Cross-Interpolation Spectral Similarity:} First, we assess whether the underlying hardware noise topology is intrinsically preserved across different resampling filters or if severe low-pass filtering scrambles the trace. Let $\tilde{P}_{m,t}$ denote the flattened average Log-PSD vector for a specific scanner model $m \in \mathcal{M}$, processed with an interpolation kernel $t \in T$. To quantify the structural preservation, we compute the pairwise Cosine Similarity across all interpolation combinations:
    \begin{equation}
        \text{Sim}(\tilde{P}_{A,t_1}, \tilde{P}_{B,t_2}) = \frac{\sum_{i=1}^{N} \tilde{P}_{A,t_1,i} \cdot \tilde{P}_{B,t_2,i}}{\sqrt{\sum_{i=1}^{N} \tilde{P}_{A,t_1,i}^2} \cdot \sqrt{\sum_{i=1}^{N} \tilde{P}_{B,t_2,i}^2}}
    \end{equation}
    Robust preservation of the KSS is indicated by high intra-scanner coherence across differing interpolators while maintaining structurally antagonistic orthogonality between different manufacturers.

    \item \textbf{Stability of the Anatomical Baseline:} Second, we evaluate the stability of the global anatomical prior under varying registration typologies. We compute a distinct global baseline $\tilde{P}_{\text{global},t}$ for each interpolation kernel $t$, utilizing exclusively the authentic training cohort corresponding to that specific preprocessing technique. By extracting and comparing the 1D radial profiles of these baselines, we mathematically quantify the exact spectral footprint and the high-frequency attenuation induced by smoother kernels relative to edge-preserving kernels.
\end{itemize}

\subsubsection{Results and Analysis: Cross-Interpolation Spectral Integrity}

\begin{figure}[t!]
    \centering
    \includegraphics[width=\columnwidth]{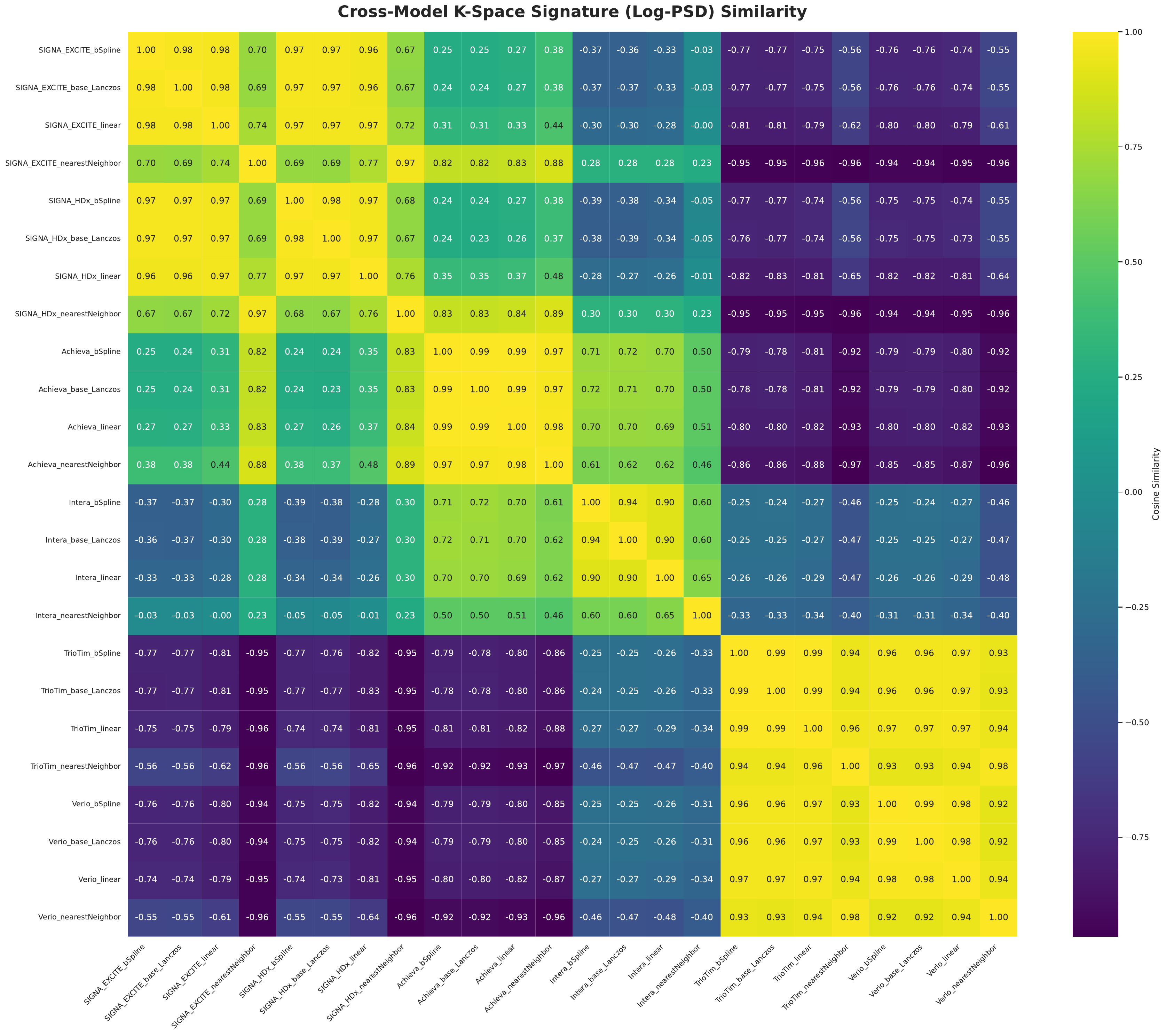}
    \caption{Cross-interpolation Cosine Similarity matrix of the isolated KSS. Continuous interpolation kernels (Lanczos, B-Spline, and Linear) demonstrate robust intra-scanner coherence and preserve the structurally antagonistic orthogonality between different manufacturers. Conversely, the Nearest Neighbor kernel introduces severe spatial aliasing, destroying the intrinsic hardware topologies and generating false cross-vendor alignments.}
    \label{fig:ablation_heatmap}
\end{figure}

The empirical results from the cross-interpolation similarity matrix (Fig.~\ref{fig:ablation_heatmap}) strongly validate the robustness of the KSS representation under standard clinical preprocessing workflows. For continuous interpolation kernels (Lanczos, B-Spline, and Linear), we observe exceptional intra-scanner coherence, with similarity scores consistently between $0.96$ and $1.00$ for volumes acquired from the same physical scanner but processed with different filters. Crucially, the structurally antagonistic orthogonality between different manufacturers is perfectly preserved. For instance, GE and Siemens hardware maintain strong negative correlations (ranging from $-0.75$ to $-0.85$) regardless of whether an edge-preserving (Lanczos) or smoothing (Linear) filter is applied. This confirms that the KSS does not rely on fragile, ultra-high-frequency micro-noise, but rather captures a macroscopic, hardware-specific spectral geometry that securely survives aggressive low-pass spatial filtering.

Conversely, the Nearest Neighbor interpolator exhibits a predictable but severe structural collapse. The similarity analysis reveals drastic intra-domain degradation and, more alarmingly, false structural alignments across physically disjoint hardware (e.g., introducing an artificial positive correlation of $>0.80$ between GE and Philips configurations). This failure occurs because the jagged, step-like boundaries inherent to nearest-neighbor resampling inject massive amounts of algorithmic high-frequency noise (spatial aliasing), which dominates the spectrum and fundamentally overwrites the subtle, intrinsic hardware traces. 

\subsubsection{Results and Analysis: Stability of the Anatomical Baseline}

\begin{figure}[t!]
\centering
    \includegraphics[width=\columnwidth]{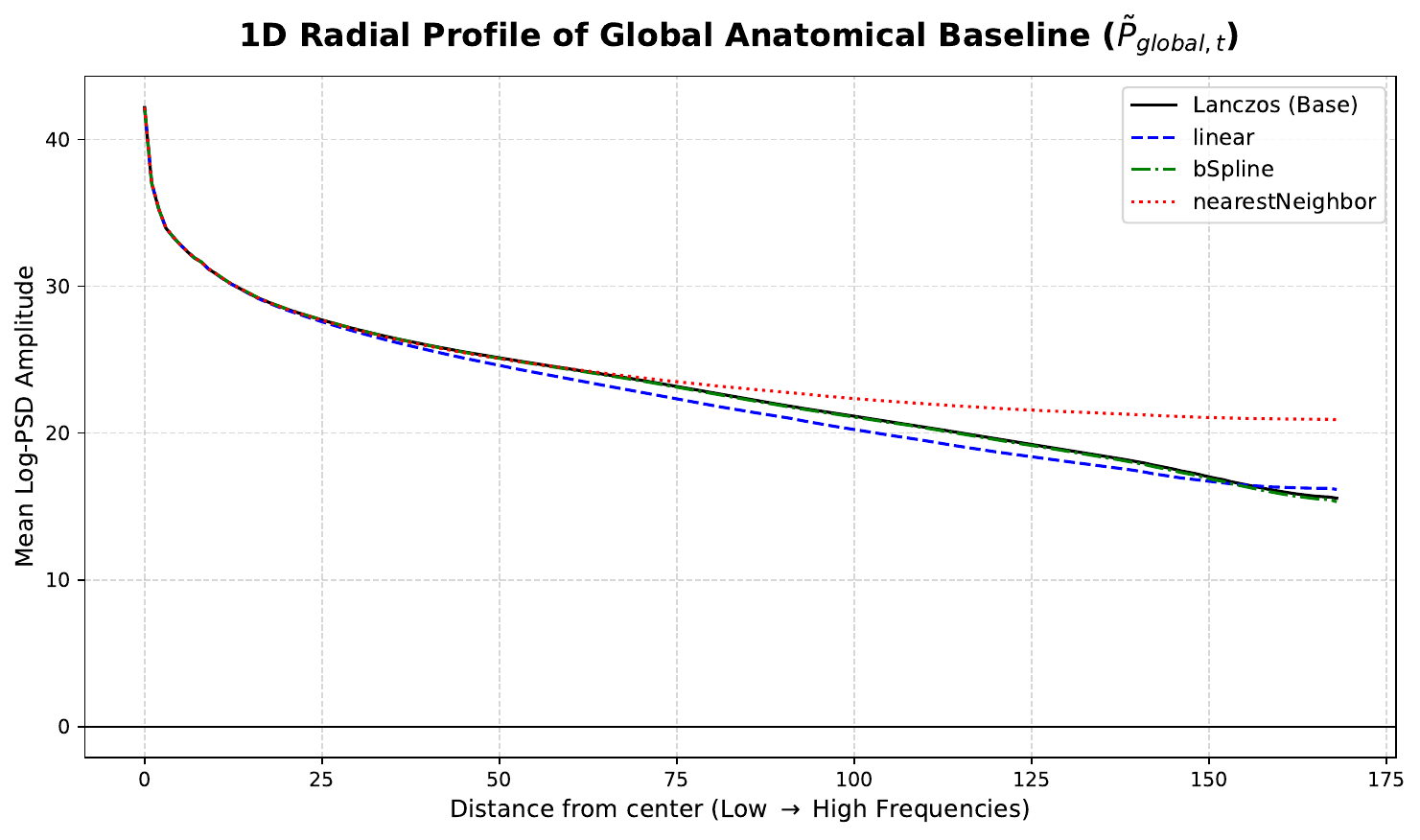}
    \caption{1D radial profiles of the global anatomical baselines ($\tilde{P}_{global, t}$) computed across different spatial resampling typologies. All kernels preserve the macroscopic anatomical structures (low frequencies). In the high-frequency tails, distinct filter footprints emerge: the Linear kernel exhibits severe low-pass attenuation, while the Nearest Neighbor kernel maintains an artificial high-energy plateau indicative of severe spatial aliasing.}
    \label{fig:ablation_radial}
\end{figure}

To mathematically contextualize the spectral behavior observed in the similarity matrix, we analyze the 1D radial profiles of the interpolation-specific global anatomical baselines, $\tilde{P}_{global,t}$ (Fig.~\ref{fig:ablation_radial}). In the low-frequency spectrum, which inherently encodes the macroscopic patient anatomy, all four interpolation typologies closely align.

As the profiles transition to the high-frequency tails, the distinct spectral footprints of the mathematical filters emerge. The baseline profiles for Lanczos and B-Spline remain tightly coupled throughout the entire frequency spectrum, indicating that the spectral footprint of our framework is highly comparable to the B-Spline registration utilized in ubiquitous neuro-clinical pipelines (e.g., FreeSurfer). However, we specifically selected the Lanczos kernel for our primary pipeline because of its precise edge-preserving properties, which are optimal for retaining the localized high-frequency hardware noise topologies required for accurate KSS extraction. The Linear interpolator profile exhibits a rapid, premature energy decay characteristic of a rigorous low-pass filter. However, because the discriminative power of the KSS is distributed globally across the spectrum, this localized high-frequency attenuation does not degrade the inter-vendor orthogonality. 

Finally, the isolated radial profile of the Nearest Neighbor interpolator mathematically explains its failure in the similarity matrix. Rather than decaying naturally, the profile maintains an unnaturally high energy plateau in the extreme high frequencies. This is the direct spectral manifestation of severe spatial aliasing: the jagged, step-like boundaries introduced by nearest-neighbor voxel copying inject a massive amount of algorithmic high-frequency noise into the K-space. This artificial energy dominates the spectrum and fundamentally overwrites the subtle, intrinsic hardware traces, confirming that KSS isolation strictly requires continuous spatial resampling kernels to function reliably.

\subsection{Extended Cross-Dataset Performance Evaluation}

\begin{table*}[t!]
    \centering
    \resizebox{\textwidth}{!}{%
    \begin{tabular}{lll ccc ccc ccc ccc ccc ccc}
        \toprule
        \multirow{3}{*}{} & \multirow{3}{*}{Train Dataset} & \multirow{3}{*}{Model} & \multicolumn{15}{c}{Test Dataset} \\
        
        & & & \multicolumn{3}{c}{MAISI} & \multicolumn{3}{c}{Med} & \multicolumn{3}{c}{$SS_S$} & \multicolumn{3}{c}{$SS_R$} & \multicolumn{3}{c}{Open Dataset} & \multicolumn{3}{c}{Avg} \\
        \cmidrule(lr){4-6} \cmidrule(lr){7-9} \cmidrule(lr){10-12} \cmidrule(lr){13-15} \cmidrule(lr){16-18} \cmidrule(lr){19-21}
        & & & AP & Acc & AUC & AP & Acc & AUC & AP & Acc & AUC & AP & Acc & AUC & AP & Acc & AUC & AP & Acc & AUC \\
        \midrule
        
    \multirow{24}{*}{\rotatebox[origin=c]{90}{MRIs}}
    & \multirow{6}{*}{MAISI}
     & ResNet3D & 1.000 & 1.000 & 1.000 & 1.000 & 1.000 & 1.000 & 0.996 & 0.522 & 0.996 & 0.997 & 0.592 & 0.997 & 0.995 & 0.580 & 0.918 & 0.998 & 0.779 & 0.998 \\
    && ShuffleNet3D & 0.681 & 0.846 & 0.893 & 0.886 & 0.845 & 0.961 & 0.620 & 0.845 & 0.859 & 0.461 & 0.594 & 0.706 & 0.967 & 0.532 & 0.608 & 0.662 & 0.782 & 0.855 \\
    && MobileNet3D & 1.000 & 0.962 & 1.000 & 1.000 & 0.962 & 1.000 & 0.668 & 0.611 & 0.879 & 0.428 & 0.586 & 0.668 & 0.978 & 0.530 & 0.678 & 0.774 & 0.780 & 0.887 \\
    && MLP-Mixer3D & 1.000 & 1.000 & 1.000 & 1.000 & 1.000 & 1.000 & 0.314 & 0.500 & 0.018 & 0.353 & 0.514 & 0.059 & 0.976 & 0.525 & 0.650 & 0.667 & 0.754 & 0.519 \\
    && MLP-Mixer-arcface3D & 1.000 & 1.000 & 1.000 & 1.000 & 1.000 & 1.000 & 0.394 & 0.500 & 0.326 & 0.449 & 0.503 & 0.351 & 0.973 & 0.527 & 0.604 & 0.711 & 0.751 & 0.669 \\
    \cmidrule(lr){2-21}
    & \multirow{6}{*}{Med}
     & ResNet3D & 0.654 & 0.499 & 0.772 & 1.000 & 1.000 & 1.000 & 0.765 & 0.500 & 0.807 & 0.883 & 0.500 & 0.897 & 0.988 & 0.288 & 0.822 & 0.826 & 0.625 & 0.869 \\
    && ShuffleNet3D & 0.606 & 0.499 & 0.707 & 1.000 & 1.000 & 1.000 & 0.900 & 0.500 & 0.914 & 0.830 & 0.500 & 0.798 & 0.967 & 0.288 & 0.598 & 0.834 & 0.625 & 0.855 \\
    && MobileNet3D & 0.453 & 0.622 & 0.692 & 1.000 & 1.000 & 1.000 & 0.489 & 0.623 & 0.733 & 0.570 & 0.623 & 0.800 & 0.983 & 0.354 & 0.621 & 0.628 & 0.717 & 0.806 \\
    && MLP-Mixer3D & 0.967 & 0.499 & 0.958 & 1.000 & 1.000 & 1.000 & 0.373 & 0.500 & 0.288 & 0.648 & 0.500 & 0.580 & 0.984 & 0.288 & 0.754 & 0.747 & 0.625 & 0.706 \\
    && MLP-Mixer-arcface3D & 0.924 & 0.499 & 0.954 & 1.000 & 1.000 & 1.000 & 0.809 & 0.500 & 0.869 & 0.807 & 0.500 & 0.840 & 0.995 & 0.288 & 0.913 & 0.885 & 0.625 & 0.916 \\
    \cmidrule(lr){2-21}
    & \multirow{6}{*}{$SS_S$}
     & ResNet3D & 0.455 & 0.499 & 0.505 & 1.000 & 1.000 & 1.000 & 1.000 & 1.000 & 1.000 & 1.000 & 1.000 & 1.000 & 0.974 & 0.745 & 0.720 & 0.864 & 0.875 & 0.876 \\
    && ShuffleNet3D & 0.976 & 0.499 & 0.981 & 1.000 & 1.000 & 1.000 & 1.000 & 1.000 & 1.000 & 1.000 & 0.894 & 1.000 & 0.970 & 0.695 & 0.689 & 0.994 & 0.848 & 0.995 \\
    && MobileNet3D & 0.324 & 0.545 & 0.463 & 0.829 & 0.922 & 0.938 & 0.830 & 0.923 & 0.938 & 0.824 & 0.923 & 0.936 & 0.967 & 0.678 & 0.793 & 0.701 & 0.828 & 0.819 \\
    && MLP-Mixer3D & 0.329 & 0.499 & 0.092 & 0.313 & 0.501 & 0.008 & 1.000 & 1.000 & 1.000 & 0.815 & 0.719 & 0.664 & 0.939 & 0.377 & 0.352 & 0.614 & 0.680 & 0.441 \\
    && MLP-Mixer-arcface3D & 0.354 & 0.499 & 0.179 & 0.461 & 0.571 & 0.192 & 1.000 & 1.000 & 1.000 & 0.818 & 0.764 & 0.670 & 0.957 & 0.430 & 0.437 & 0.658 & 0.708 & 0.510 \\
    \cmidrule(lr){2-21}
    & \multirow{6}{*}{$SS_R$}
     & ResNet3D & 0.531 & 0.499 & 0.639 & 1.000 & 1.000 & 1.000 & 1.000 & 1.000 & 1.000 & 1.000 & 1.000 & 1.000 & 0.972 & 0.745 & 0.725 & 0.883 & 0.875 & 0.910 \\
    && ShuffleNet3D & 0.962 & 0.499 & 0.979 & 1.000 & 1.000 & 1.000 & 1.000 & 1.000 & 1.000 & 1.000 & 1.000 & 1.000 & 0.978 & 0.745 & 0.751 & 0.991 & 0.875 & 0.995 \\
    && MobileNet3D & 0.619 & 0.545 & 0.860 & 0.836 & 0.922 & 0.941 & 0.645 & 0.923 & 0.879 & 0.762 & 0.923 & 0.913 & 0.976 & 0.740 & 0.745 & 0.715 & 0.828 & 0.899 \\
    && MLP-Mixer3D & 0.920 & 0.648 & 0.894 & 0.998 & 0.813 & 0.999 & 1.000 & 1.000 & 1.000 & 1.000 & 1.000 & 1.000 & 0.974 & 0.728 & 0.744 & 0.980 & 0.865 & 0.973 \\
    && MLP-Mixer-arcface3D & 0.817 & 0.582 & 0.826 & 0.408 & 0.532 & 0.134 & 1.000 & 0.992 & 1.000 & 0.999 & 0.981 & 0.999 & 0.964 & 0.566 & 0.595 & 0.806 & 0.772 & 0.740 \\
    \midrule
    
    \multirow{24}{*}{\rotatebox[origin=c]{90}{K-Space}}
    & \multirow{6}{*}{MAISI}
     & ResNet3D & 1.000 & 0.997 & 1.000 & 1.000 & 0.997 & 1.000 & 1.000 & 0.997 & 1.000 & 0.998 & 0.928 & 0.998 & 0.977 & 0.942 & 0.785 & 0.999 & 0.980 & 1.000 \\
    && ShuffleNet3D & 1.000 & 1.000 & 1.000 & 1.000 & 1.000 & 1.000 & 1.000 & 1.000 & 1.000 & 1.000 & 0.878 & 1.000 & 0.975 & 0.921 & 0.770 & 1.000 & 0.969 & 1.000 \\
    && MobileNet3D & 1.000 & 0.997 & 1.000 & 1.000 & 0.997 & 1.000 & 1.000 & 0.997 & 1.000 & 1.000 & 0.989 & 1.000 & 0.977 & 0.971 & 0.785 & 1.000 & 0.995 & 1.000 \\
    && MLP-Mixer3D & 1.000 & 0.997 & 1.000 & 1.000 & 0.997 & 1.000 & 1.000 & 0.997 & 1.000 & 0.997 & 0.889 & 0.998 & 0.948 & 0.927 & 0.684 & 0.999 & 0.970 & 0.999 \\
    && MLP-Mixer-arcface3D & 1.000 & 0.964 & 1.000 & 0.949 & 0.961 & 0.978 & 1.000 & 0.964 & 1.000 & 0.998 & 0.964 & 0.999 & 0.945 & 0.965 & 0.635 & 0.987 & 0.963 & 0.994 \\
    \cmidrule(lr){2-21}
    & \multirow{6}{*}{Med}
     & ResNet3D & 1.000 & 0.499 & 1.000 & 1.000 & 1.000 & 1.000 & 1.000 & 0.500 & 1.000 & 0.999 & 0.500 & 0.999 & 0.979 & 0.288 & 0.768 & 1.000 & 0.625 & 1.000 \\
    && ShuffleNet3D & 1.000 & 0.499 & 1.000 & 1.000 & 1.000 & 1.000 & 1.000 & 0.500 & 1.000 & 1.000 & 0.500 & 1.000 & 0.978 & 0.288 & 0.765 & 1.000 & 0.625 & 1.000 \\
    && MobileNet3D & 0.795 & 0.499 & 0.861 & 1.000 & 1.000 & 1.000 & 0.997 & 0.500 & 0.997 & 0.947 & 0.500 & 0.955 & 0.975 & 0.288 & 0.706 & 0.935 & 0.625 & 0.953 \\
    && MLP-Mixer3D & 1.000 & 0.499 & 1.000 & 1.000 & 1.000 & 1.000 & 1.000 & 0.500 & 1.000 & 0.999 & 0.500 & 0.999 & 0.979 & 0.288 & 0.767 & 1.000 & 0.625 & 1.000 \\
    && MLP-Mixer-arcface3D & 1.000 & 0.499 & 1.000 & 1.000 & 1.000 & 1.000 & 1.000 & 0.500 & 1.000 & 1.000 & 0.500 & 1.000 & 0.979 & 0.288 & 0.768 & 1.000 & 0.625 & 1.000 \\
    \cmidrule(lr){2-21}
    & \multirow{6}{*}{$SS_S$}
     & ResNet3D & 1.000 & 0.718 & 1.000 & 1.000 & 1.000 & 1.000 & 1.000 & 1.000 & 1.000 & 0.998 & 0.797 & 0.998 & 0.965 & 0.753 & 0.726 & 1.000 & 0.879 & 1.000 \\
    && ShuffleNet3D & 1.000 & 0.510 & 1.000 & 1.000 & 1.000 & 1.000 & 1.000 & 1.000 & 1.000 & 1.000 & 0.742 & 1.000 & 0.967 & 0.628 & 0.738 & 1.000 & 0.813 & 1.000 \\
    && MobileNet3D & 0.996 & 0.499 & 0.997 & 0.923 & 0.501 & 0.978 & 1.000 & 1.000 & 1.000 & 0.994 & 0.747 & 0.994 & 0.946 & 0.390 & 0.645 & 0.978 & 0.687 & 0.992 \\
    && MLP-Mixer3D & 1.000 & 0.995 & 1.000 & 0.984 & 0.989 & 0.994 & 1.000 & 0.994 & 1.000 & 0.999 & 0.994 & 0.999 & 0.948 & 0.972 & 0.677 & 0.996 & 0.993 & 0.998 \\
    && MLP-Mixer-arcface3D & 1.000 & 0.729 & 1.000 & 0.997 & 0.724 & 0.994 & 1.000 & 0.728 & 1.000 & 0.999 & 0.728 & 0.999 & 0.978 & 0.960 & 0.766 & 0.999 & 0.727 & 0.998 \\
    \cmidrule(lr){2-21}
    & \multirow{6}{*}{$SS_R$}
     & ResNet3D & 1.000 & 1.000 & 1.000 & 1.000 & 1.000 & 1.000 & 1.000 & 1.000 & 1.000 & 1.000 & 1.000 & 1.000 & 0.975 & 0.979 & 0.777 & 1.000 & 1.000 & 1.000 \\
    && ShuffleNet3D & 1.000 & 0.981 & 1.000 & 0.384 & 0.504 & 0.315 & 1.000 & 1.000 & 1.000 & 1.000 & 1.000 & 1.000 & 0.942 & 0.736 & 0.532 & 0.846 & 0.871 & 0.829 \\
    && MobileNet3D & 1.000 & 0.549 & 1.000 & 0.311 & 0.501 & 0.018 & 1.000 & 1.000 & 1.000 & 1.000 & 1.000 & 1.000 & 0.969 & 0.533 & 0.617 & 0.828 & 0.763 & 0.754 \\
    && MLP-Mixer3D & 1.000 & 0.995 & 1.000 & 0.973 & 0.994 & 0.994 & 0.974 & 0.994 & 0.994 & 0.998 & 0.994 & 0.998 & 0.994 & 0.978 & 0.905 & 0.986 & 0.994 & 0.997 \\
    && MLP-Mixer-arcface3D & 1.000 & 0.665 & 1.000 & 1.000 & 0.663 & 1.000 & 1.000 & 0.664 & 1.000 & 0.999 & 0.664 & 0.999 & 0.974 & 0.952 & 0.752 & 1.000 & 0.664 & 1.000 \\
    \midrule
    
    \multirow{24}{*}{\rotatebox[origin=c]{90}{KSS}}
    & \multirow{6}{*}{MAISI}
     & ResNet3D & 1.000 & 0.995 & 1.000 & 1.000 & 0.994 & 1.000 & 1.000 & 0.994 & 1.000 & 0.997 & 0.908 & 0.997 & 0.976 & 0.940 & 0.782 & 0.999 & 0.973 & 0.999 \\
    && ShuffleNet3D & 1.000 & 0.997 & 1.000 & 1.000 & 0.997 & 1.000 & 1.000 & 0.997 & 1.000 & 0.999 & 0.944 & 0.999 & 0.977 & 0.956 & 0.791 & 1.000 & 0.984 & 1.000 \\
    && MobileNet3D & 1.000 & 0.997 & 1.000 & 1.000 & 0.997 & 1.000 & 1.000 & 0.997 & 1.000 & 0.996 & 0.867 & 0.997 & 0.979 & 0.922 & 0.762 & 0.999 & 0.965 & 0.999 \\
    && MLP-Mixer3D & 1.000 & 1.000 & 1.000 & 1.000 & 1.000 & 1.000 & 1.000 & 1.000 & 1.000 & 1.000 & 0.972 & 1.000 & 0.979 & 0.967 & 0.767 & 1.000 & 0.993 & 1.000 \\
    && MLP-Mixer-arcface3D & 1.000 & 0.995 & 1.000 & 1.000 & 0.994 & 1.000 & 1.000 & 0.994 & 1.000 & 1.000 & 0.994 & 1.000 & 0.979 & 0.973 & 0.769 & 1.000 & 0.994 & 1.000 \\
    \cmidrule(lr){2-21}
    & \multirow{6}{*}{Med}
     & ResNet3D & 1.000 & 0.499 & 1.000 & 1.000 & 1.000 & 1.000 & 1.000 & 0.500 & 1.000 & 1.000 & 0.500 & 1.000 & 0.978 & 0.288 & 0.766 & 1.000 & 0.625 & 1.000 \\
    && ShuffleNet3D & 1.000 & 0.499 & 1.000 & 1.000 & 1.000 & 1.000 & 1.000 & 0.500 & 1.000 & 1.000 & 0.500 & 1.000 & 0.979 & 0.288 & 0.767 & 1.000 & 0.625 & 1.000 \\
    && MobileNet3D & 0.923 & 0.499 & 0.943 & 1.000 & 1.000 & 1.000 & 1.000 & 0.500 & 1.000 & 0.993 & 0.500 & 0.993 & 0.977 & 0.288 & 0.746 & 0.979 & 0.625 & 0.984 \\
    && MLP-Mixer3D & 1.000 & 0.759 & 1.000 & 1.000 & 1.000 & 1.000 & 1.000 & 1.000 & 1.000 & 1.000 & 0.942 & 1.000 & 0.979 & 0.842 & 0.770 & 1.000 & 0.925 & 1.000 \\
    && MLP-Mixer-arcface3D & 1.000 & 1.000 & 1.000 & 1.000 & 1.000 & 1.000 & 1.000 & 1.000 & 1.000 & 1.000 & 1.000 & 1.000 & 0.979 & 0.983 & 0.769 & 1.000 & 1.000 & 1.000 \\
    \cmidrule(lr){2-21}
    & \multirow{6}{*}{$SS_S$}
     & ResNet3D & 1.000 & 0.499 & 1.000 & 1.000 & 1.000 & 1.000 & 1.000 & 1.000 & 1.000 & 1.000 & 0.631 & 1.000 & 0.966 & 0.570 & 0.735 & 1.000 & 0.782 & 1.000 \\
    && ShuffleNet3D & 0.999 & 0.499 & 0.999 & 0.310 & 0.501 & 0.000 & 1.000 & 1.000 & 1.000 & 0.999 & 0.681 & 0.999 & 0.939 & 0.358 & 0.521 & 0.827 & 0.670 & 0.750 \\
    && MobileNet3D & 1.000 & 0.992 & 1.000 & 1.000 & 0.997 & 1.000 & 1.000 & 0.997 & 1.000 & 0.993 & 0.861 & 0.993 & 0.950 & 0.917 & 0.693 & 0.998 & 0.962 & 0.998 \\
    && MLP-Mixer3D & 1.000 & 0.997 & 1.000 & 1.000 & 0.997 & 1.000 & 1.000 & 0.997 & 1.000 & 1.000 & 0.997 & 1.000 & 0.979 & 0.980 & 0.770 & 1.000 & 0.997 & 1.000 \\
    && MLP-Mixer-arcface3D & 1.000 & 0.997 & 1.000 & 1.000 & 0.997 & 1.000 & 1.000 & 0.997 & 1.000 & 1.000 & 0.997 & 1.000 & 0.979 & 0.981 & 0.770 & 1.000 & 0.997 & 1.000 \\
    \cmidrule(lr){2-21}
    & \multirow{6}{*}{$SS_R$}
     & ResNet3D & 1.000 & 1.000 & 1.000 & 1.000 & 0.994 & 1.000 & 1.000 & 1.000 & 1.000 & 1.000 & 1.000 & 1.000 & 0.988 & 0.981 & 0.834 & 1.000 & 0.999 & 1.000 \\
    && ShuffleNet3D & 1.000 & 1.000 & 1.000 & 1.000 & 1.000 & 1.000 & 1.000 & 1.000 & 1.000 & 1.000 & 1.000 & 1.000 & 0.974 & 0.979 & 0.771 & 1.000 & 1.000 & 1.000 \\
    && MobileNet3D & 1.000 & 0.997 & 1.000 & 1.000 & 1.000 & 1.000 & 1.000 & 1.000 & 1.000 & 1.000 & 1.000 & 1.000 & 0.951 & 0.976 & 0.691 & 1.000 & 0.999 & 1.000 \\
    && MLP-Mixer3D & 0.991 & 0.967 & 0.995 & 1.000 & 0.997 & 1.000 & 1.000 & 0.997 & 1.000 & 1.000 & 0.997 & 1.000 & 0.986 & 0.968 & 0.817 & 0.998 & 0.990 & 0.999 \\
    && MLP-Mixer-arcface3D & 1.000 & 0.997 & 1.000 & 1.000 & 0.997 & 1.000 & 1.000 & 0.997 & 1.000 & 1.000 & 0.997 & 1.000 & 0.979 & 0.980 & 0.769 & 1.000 & 0.997 & 1.000 \\
    \bottomrule
    \end{tabular}
    } 
    \caption{Comprehensive cross-dataset performance evaluation. Each model is trained exclusively on a single source dataset (rows) and subsequently evaluated across all available target datasets (columns).}
    \label{tab:cross_dataset_results_split}
\end{table*}

As referenced in the main manuscript, Figure~\ref{res:resuls} visualizes the global accuracy trends across different architectural configurations and spatial domains. Table~\ref{tab:cross_dataset_results_split} in this supplementary material provides the exhaustive, granular metrics that complement this visual summary, presenting a comprehensive multi-domain cross-dataset evaluation.

\subsubsection{Evaluation Setup and Metrics}
To comprehensively test zero-shot generalization, models were trained exclusively on a single generative dataset (MAISI, Med, SS\_S, SS\_R) and evaluated across all available datasets, including a strictly isolated Open Dataset. The exhaustive results of this analysis, which spans three distinct representation spaces, are detailed in Table~\ref{tab:cross_dataset_results_split}:
\begin{itemize}
    \item The raw spatial Representative Space.
    \item The standard complex K-Space.
    \item Our KSS domain.
\end{itemize}
To provide a holistic view of model robustness, performance is quantified across three metrics: Average Precision (AP), Accuracy (Acc), and ROC AUC.

\subsubsection{Architectural Robustness and Domain Generalization}
The tabular data explicitly corroborates the Out-of-Distribution (OOD) trends highlighted in Figure~\ref{res:resuls} of the main text. Specifically, the extended metrics reveal the following core insights regarding architectural inductive biases:

\begin{itemize}
    \item \textbf{CNN Vulnerability to Domain Shifts:} Traditional 3D Convolutional Neural Networks (ResNet3D, ShuffleNet3D, MobileNet3D) exhibit severe vulnerability when subjected to cross-generator domain shifts. As shown in Table~\ref{tab:cross_dataset_results_split}, when operating in the Representative Space and standard K-Space domains, these CNN backbones frequently experience drastic accuracy and AP degradation when evaluated on the unseen Open Dataset.
    \item \textbf{Superior Stability of KSS and MLP-Mixer:} Conversely, the MLP-Mixer variants, and particularly the proposed \textit{MLP-Mixer-arcface3D}, demonstrate exceptional generalization stability. When coupled with the KSS representation space, the network successfully leverages its global receptive field to process diffuse spectral anomalies distributed across the entire spectrum. 
\end{itemize}

As detailed throughout Table~\ref{tab:cross_dataset_results_split}, the KSS paired with the ArcFace metric-learning head effectively minimizes the performance gap between the closed validation sets (seen generators) and the OOD Open Dataset (unseen scanners). By maintaining high accuracy and ROC AUC scores across varying generative pipelines, this comprehensive breakdown empirically confirms the robustness of the KSS framework for multi-scanner clinical deployment.

\subsection{Zero-Shot Domain Adaptation}
\begin{table*}[ht!]
    \centering
    \resizebox{\textwidth}{!}{%
    \begin{tabular}{llccccc}
        \toprule
        \textbf{Unseen Data} & \textbf{Modality} & \textbf{Accuracy (\%)} & \textbf{Macro $F_1$ (\%)} & \textbf{ROC AUC (\%)} & \textbf{Sensitivity (\%)} & \textbf{Specificity (\%)} \\
        \midrule
        \multirow{2}{*}{GE}      & Baseline (Raw) & 59.33 & 37.24 & 76.51 & 0.00  & \textbf{100.00} \\
        & \textbf{KSS-Harmonized} & \textbf{68.39} & \textbf{58.56} & \textbf{89.98} & \textbf{24.20}  & 98.69 \\
        \midrule
        \multirow{2}{*}{Siemens} & Baseline (Raw) & 70.93 & 70.41 & 85.05 & 90.32 & 53.99  \\
        & \textbf{KSS-Harmonized} & \textbf{78.70} & \textbf{78.58} & \textbf{90.95} & \textbf{92.47} & \textbf{66.67}  \\
        \midrule
        \multirow{2}{*}{Philips} & Baseline (Raw) & \textbf{60.05} & \textbf{56.46} & \textbf{71.84} & \textbf{28.30} & \textbf{99.42}  \\
        & \textbf{KSS-Harmonized} & 44.91 & 31.78 & 62.03 & 0.94 & \textbf{99.42}  \\
        \bottomrule
    \end{tabular}
    }
    \caption{Comparative Zero-Shot Domain Adaptation Results for Gender Classification. Evaluation of the Baseline (Raw Spatial) approach versus the Proposed KSS-Harmonized framework across three Leave-One-Domain-Out splits. Bold text highlights the top-performing method within each independent testing split.}
    \label{tab:kss_results}
\end{table*}
To rigorously evaluate the generalization capabilities of our proposed framework against hardware-induced domain shifts, we frame the evaluation as a Multi-Source Domain Adaptation (MSDA) problem. We adopt a Leave-One-Domain-Out validation strategy across three distinct scanner manufacturers (GE, Philips, Siemens). For each split, one manufacturer is strictly isolated to form the unseen test set, while the remaining two form the source domains ($\mathcal{D}_{train}$). To prevent implicit memorization, we enforce a strict patient-level split across all subsets.
A critical challenge in this MSDA setting is defining the anatomical baseline without exposing the test distribution to the model. Let $\mathcal{M} = \{M_1, M_2, M_3\}$ denote the complete set of available MRI manufacturers. In our Leave-One-Domain-Out strategy, one manufacturer $M_{test} \in \mathcal{M}$ is strictly isolated as the unseen target domain. Therefore, the global anatomical baseline ($\tilde{P}_{global}$) is computed dynamically and exclusively on the training subset of the remaining source domains, defined as $\mathcal{M}_{source} = \mathcal{M} \setminus \{M_{test}\}$, where $|\mathcal{M}_{source}| = 2$. For each source domain manufacturer $M_j \in \mathcal{M}_{source}$, we extract the isolated trace $KSS_{M_j}$ by mapping the mean complex K-Space of that specific manufacturer into the Log-PSD domain and subtracting this global baseline.
We compare our method against a Baseline Setup trained directly on raw spatial MRI volumes for the downstream task of gender classification. In our Proposed Setup, the network is trained on KSS-Filtered images. Because the source-specific traces are defined in the Log-PSD domain, the structural subtraction must occur within this space to be mathematically valid. Specifically, for a raw spatial volume $V_{raw}$ with a complex frequency representation $\hat{V}_{raw} = \mathcal{F}(V_{raw})$ and corresponding phase $\angle \hat{V}_{raw}$, we compute its Log-PSD $\tilde{P}_{raw}$ and subtract the trace:

\begin{equation}
\tilde{P}_{filtered} = \tilde{P}_{raw} - KSS_{M_j}
\end{equation}

The filtered spatial volume is then reconstructed by inverting this modified Log-PSD back to a linear magnitude representation, recombining it with the original complex phase, and applying the Inverse 3D DFT to extract the real spatial component:

\begin{equation}
V_{filtered} = \Re\left( \mathcal{F}^{-1}\left( \sqrt{\max(0, e^{\tilde{P}_{filtered}} - 1)} \cdot e^{j \angle \hat{V}_{raw}} \right) \right)
\end{equation}

Crucially, during inference, the unseen target domain images are processed in their native spatial form (without KSS subtraction) to strictly evaluate zero-shot adaptation.

\subsubsection{Results and Domain Shift Analysis}

The comparative results for this zero-shot evaluation are reported in Table \ref{tab:kss_results}. The empirical findings demonstrate that the KSS-Harmonized representation yields consistent generalization improvements for the GE and Siemens unseen testing domains. Specifically, for Siemens, the proposed method achieves an absolute increase of 7.77\% in Accuracy and 8.17\% in Macro $F_1$ score relative to the raw baseline. The KSS approach also balances the trade-off between sensitivity and specificity more effectively, substantially increasing specificity for Siemens (from 53.99\% to 66.67\%) while mildly improving sensitivity (from 90.32\% to 92.47\%). Furthermore, for GE, the framework delivers a significant 9.06\% improvement in Accuracy (overcoming the baseline's majority-class default of 59.33\% to reach 68.39\%) and a 21.32\% enhancement in Macro $F_1$. These improvements strongly validate our core hypothesis: subtracting hardware-specific frequency components successfully suppresses confounding scanner artifacts, forcing the network to prioritize invariant, high-level anatomical features.
However, Philips highlights a critical boundary condition. The KSS model experiences a severe performance degradation on this domain, with Accuracy falling to 44.91\% and Sensitivity collapsing to 0.94\% (compared to the baseline's 60.05\% and 28.30\%, respectively). This degradation can be directly contextualized through our earlier KSS similarity analysis. When Philips is isolated as the target domain, the network is trained exclusively on SIEMENS and GE datasets. As proven in previous sections, certain Philips models exhibit extreme orthogonality to these training domains, with negative correlations to Siemens reaching up to -0.79. Because the unseen Philips volumes are processed in their native form during zero-shot inference, this unmitigated, structurally antagonistic hardware noise introduces a highly disruptive domain shift that the network fails to bridge. This confirms that strictly disjoint noise topologies represent a hard limit for current zero-shot spatial adaptation techniques.

\subsection{Visual Explainability of the KSS}

\begin{figure*}[t!]
    \centering
    \includegraphics[width=\textwidth]{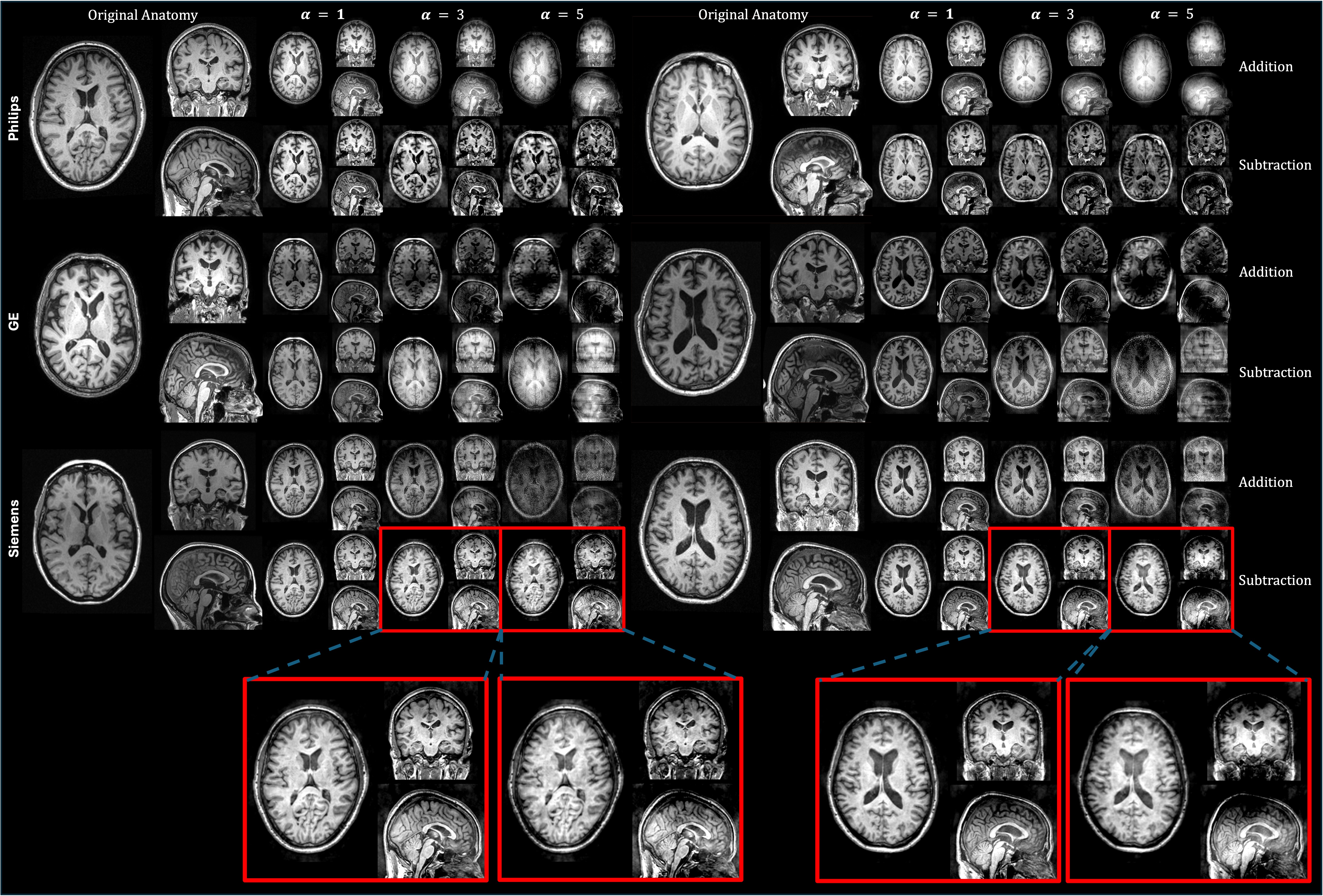}
    \caption{Visual explainability and feature disentanglement of the KSS. The figure is organized by manufacturer: the first row displays Philips images, providing one example for each of its scanner models; the second and third rows similarly present examples for GE and Siemens scanners, respectively. The left panel displays the Original Anatomy in a triplanar view (axial, coronal, sagittal). The subsequent columns illustrate the spatial reconstruction after modulating the KSS within the logarithmic frequency domain ($\tilde{P}_{mod} = \tilde{P}_{orig} \pm \alpha \cdot \text{KSS}$) at increasing intensities ($\alpha \in \{1, 3, 5\}$). Addition ($\tilde{P}_{orig} + \alpha \cdot \text{KSS}$) amplifies the intrinsic hardware noise footprint, exacerbating the scanner-specific spatial contrast. Conversely, Subtraction ($\tilde{P}_{orig} - \alpha \cdot \text{KSS}$) acts as an aggressive spectral notch filter, suppressing trace frequencies and introducing severe ringing artifacts at high $\alpha$ values. The red-bordered insets at the bottom provide magnified views of select reconstructions, allowing for a more detailed examination of the fine-grained visual impacts caused by these modulations. Across all extreme manipulations, the macroscopic patient anatomy remains structurally intact, empirically demonstrating that the KSS effectively isolates scanner-specific acquisition traces without relying on underlying physiological structures.}
    \label{fig:kss_explainability}
\end{figure*}
To empirically demonstrate that the proposed KSS successfully disentangles hardware-specific acquisition traces from patient anatomy, we introduce a visual explainability experiment. While our detection framework isolates specific hardware and algorithmic traces by operating entirely in the high-dimensional frequency domain, visual interpretation of these spectral tensors remains unintuitive for clinical verification. Therefore, we formalize a reverse-mapping protocol to project the isolated KSS footprint back into the representative spatial domain.

\subsubsection{Mathematical Formulation}
Building upon the reconstruction formulation established for the Zero-Shot KSS-Filtering, we introduce a scalar multiplier $\alpha$ to dynamically modulate the intensity of the isolated KSS. For an authentic spatial MRI volume, we extract its original Log-PSD, denoted as $\tilde{P}_{orig} = \ln(1 + |\hat{V}|^2)$. We then apply the modulation directly in this logarithmic domain:

\begin{equation}
\tilde{P}_{mod} = \tilde{P}_{orig} \pm \alpha \cdot KSS
\end{equation}

This formulation generalizes the prior filtering process: while our zero-shot adaptation utilizes subtraction with $\alpha = 1$ to neutralize the hardware trace, varying $\alpha$ and the operation sign allows us to artificially amplify or suppress the trace's footprint. The modified tensor is then inverted back to a linear magnitude representation ($|\hat{V}_{mod}| = \sqrt{e^{\tilde{P}_{mod}} - 1}$), recombined with the original complex phase information, and transformed back into the spatial domain via the Inverse 3D DFT.

\subsubsection{Empirical Observations and Feature Disentanglement}

By varying the scalar $\alpha$ and the operation sign, we can artificially force the isolated hardware-specific frequencies to manifest visually within the native spatial volume, as illustrated in Fig.~\ref{fig:kss_explainability}. We specifically evaluate extreme conditions (e.g., $\alpha = 3$) to analyze the behavioral boundaries of the isolated features. To facilitate a more granular examination, the red-bordered insets in Fig.~\ref{fig:kss_explainability} provide magnified views of the resulting reconstructions.

\begin{itemize}
\item \textbf{Trace Subtraction (Targeted Frequency Suppression):} When applying an aggressive subtraction ($\tilde{P}_{orig} - \alpha \cdot \text{KSS}$), the operation acts mathematically as a highly specific, multi-band notch filter. By aggressively zeroing out the spectral energy localized within the KSS topology, the corresponding spatial reconstruction suffers from severe ringing artifacts and structural degradation. As highlighted in the magnified insets, these artifacts are highly visible in the background noise outside the skull. This confirms that the KSS precisely targets the frequencies responsible for stable spatial edge reconstruction and hardware noise.

\item \textbf{Trace Addition (Hardware Footprint Amplification):} Conversely, the additive operation ($\tilde{P}_{orig} + \alpha \cdot \text{KSS}$) exponentially amplifies the intrinsic physical noise traces left by the sensing device. Rather than destroying the image structure, this injection exacerbates the native spatial contrast and amplifies the background noise profile consistently with the original scanner's topology, a structural shift that is distinctly visible within the red-bordered magnified regions.
\end{itemize}

Crucially, under both extreme manipulations, the macroscopic patient anatomy (e.g., ventricles, sulci, and cortical boundaries) remains structurally unmoved and intact. This visual feature disentanglement provides empirical support for our core hypothesis: the KSS successfully isolates scanner-specific acquisition traces and algorithmic generation traces without relying on patient-specific anatomical shortcuts. This ensures that the downstream neural networks base their medical deepfake detection strictly on hardware and generative artifacts.

\section*{Acknowledgments}
\textit{Data collection and sharing for the Alzheimer's Disease Neuroimaging Initiative (ADNI) is funded by the National Institute on Aging (National Institutes of Health Grant U19AG024904). The grantee organization is the Northern California Institute for Research and Education. In the past, ADNI has also received funding from the National Institute of Biomedical Imaging and Bioengineering, the Canadian Institutes of Health Research, and private sector contributions through the Foundation for the National Institutes of  Health (FNIH) including generous contributions from the following: AbbVie, Alzheimer’s Association; Alzheimer’s Drug Discovery Foundation; Araclon Biotech; BioClinica, Inc.; Biogen; Bristol-Myers Squibb Company; CereSpir, Inc.; Cogstate; Eisai Inc.; Elan Pharmaceuticals, Inc.; Eli Lilly and Company; EuroImmun; F. Hoffmann-La Roche Ltd and its affiliated company Genentech, Inc.; Fujirebio; GE Healthcare; IXICO Ltd.; Janssen Alzheimer Immunotherapy Research \& Development, LLC.; Johnson \& Johnson Pharmaceutical Research \& Development LLC.; Lumosity; Lundbeck; Merck \& Co., Inc.; Meso Scale Diagnostics, LLC.; NeuroRx Research; Neurotrack Technologies; Novartis Pharmaceuticals Corporation; Pfizer Inc.; Piramal Imaging; Servier; Takeda Pharmaceutical Company; and Transition Therapeutics} \newline
\textit{Data used in the preparation of this article was obtained on 2025-05-20 from the Parkinson’s
Progression Markers Initiative (PPMI) database, RRID: SCR\_006431. For up-to-date information on the study, visit www.ppmi-info.org. 
PPMI – a public-private partnership – is funded by the Michael J. Fox Foundation for Parkinson’s Research and funding partners (full list available at https://www.ppmi-info.org/about-ppmi/who-we-are/study-sponsors).}

\section*{Dataset statements}\label{sec:dataset-statements}
\textit{Data used in the preparation of this article were obtained from the Alzheimer's Disease Neuroimaging Initiative (ADNI) database (adni.loni.usc.edu). The ADNI was launched in 2003 as a public-private partnership, led by Principal Investigator Michael W. Weiner, MD. The original goal of ADNI was to test whether serial MRI, positron emission tomography (PET), other biological markers, and clinical and neuropsychological assessment can be combined to measure the progression of MCI and early AD. The current goals include validating biomarkers for clinical trials, improving the generalizability of ADNI data by increasing diversity in the participant cohort, and to provide data concerning the diagnosis and progression of AD to the scientific community. For up-to-date information, see adni.loni.usc.edu.}

\end{document}